\documentclass[runningheads,12pt]{llncs}
\usepackage[T1]{fontenc}
\usepackage{graphicx}
\usepackage{placeins}
\usepackage[hidelinks]{hyperref}
\usepackage{url}
\usepackage{booktabs} 
\usepackage{subcaption}
\usepackage{graphicx}
\usepackage{multirow}
\usepackage{array}
\usepackage{ragged2e}
\usepackage[table]{xcolor}
\usepackage{amsmath}
\usepackage{wrapfig}

\usepackage{geometry}

\begin{document}
	
	
	
	%
	
	\title{Enhancing Ultra-Low-Bit Quantization of Large Language Models Through Saliency-Aware Partial Retraining}
	\titlerunning{Saliency-Aware Partial Retraining for Ultra-Low-Bit Quantization}
	%
	\author{Deyu Cao\inst{1}\orcidID{0009-0003-8563-3842} \and
		Samin Aref\inst{2*}\orcidID{0000-0002-5870-9253}}
	\authorrunning{D. Cao and S. Aref}
	%
	\institute{The University of Tokyo, 7-3-1 Hongo, Bunkyo-ku, Tokyo 113-8654, Japan \and
		Department of Mechanical and Industrial Engineering, University of Toronto,\\ 5 King's College Rd, Toronto, ON M5S 3G8, Canada\\
		*Corresponding author's email: \email{s.aref@utoronto.ca}
	}
	\maketitle       
	\begin{abstract}
		\normalsize
		
		The growing use of large language models has raised environmental and economic concerns about their intensity of resource usage during inference. Serving these models to each user requires substantial energy and water for cooling. Model compression techniques like quantization can shrink large language models and make them more resource efficient at the cost of potential performance degradation. Quantization methods compress model size through replacing their high-precision parameters by quantized values of lower precision. Among existing methods, the ApiQ method achieves superior accuracy preservation at minimal memory and time overhead. We investigate two ideas to extend performance in ultra-low-bit quantization beyond ApiQ’s level. First, we look into combining existing quantization-aware training techniques with ApiQ’s partial training. We show that this does not outperform the baseline ApiQ method with limited training data and frozen weights. This leads to two key insights: (1) The substantial representational capacity that is gained through full retraining is unlikely to be feasible through partial training. (2) This gain may depend on using a large and diverse dataset in quantization-aware training. Second, through a novel approach informed by the two insights, we propose an ultra-low-bit quantization method that builds upon ApiQ and extends its performance without the need for full retraining. This publicly available method relies on a saliency-aware regularization term that prioritizes preserving the most impactful parameters during quantization. Our experiments on LLaMA 7B and 13B benchmarks demonstrate that our method reduces the ApiQ’s accuracy degradation by 10.85\% and 7.54\% respectively. A Python implementation of the proposed quantization method is publicly available on GitHub \url{https://github.com/TokuyuSou/ULB-SAPR}. \\
		
		\small		
		This is a post-peer-review accepted manuscript from the proceedings of the 22nd International Conference on Modeling Decisions for Artificial Intelligence (MDAI'25). The publisher authenticated version (version of record) and full citation details are available on Springer's website (Lecture Notes in Artificial Intelligence 15957). DOI: \url{https://doi.org/10.1007/978-3-032-00891-6_28}
				

		
		
		
		\normalsize
	\end{abstract}
	%
	%
	%
	
	\section{Introduction}
	\label{sec:intro}
	Large Language Models (LLMs) have substantially advanced the field of artificial intelligence and specifically the area of natural language processing. However, their substantial memory and computational requirements have led to economic and environmental concerns raised about their intensive use of resources. Moreover, there are technical challenges for deploying LLMs on resource-constrained devices. Model compression techniques, including quantization, shrink LLMs in different ways and allow a more resource-efficient deployment \cite{zhu2024survey}. Quantization refers to methods that reduce model size by decreasing the bit precision of model parameters \cite{shen2024exploring,lee2024owq}. The reduced model size, in turn, improves inference speed, which is especially useful for deploying models on edge devices \cite{lang2024comprehensive,jin2024comprehensive}. The application of different quantization methods on LLMs has been widely studied in recent review papers \cite{lang2024comprehensive,jin2024comprehensive}. 
	
	Quantization involves mapping high-precision values (typically in float32 or float16) to lower-precision representations (such as int8, int4, or even fewer bits). For instance, when quantizing to 8-bit integers, each parameter can take one of $2^8$ possible discrete values. The specific formulation of a quantization method determines how this mapping is performed. In Appendix \ref{sec:rtn_example}, we present the mathematical formulation and a numerical example of the simplest quantization method, known as Round-To-Nearest quantization to provide technical details on concepts such as scaling factors, clipping thresholds, and zero points.
	
	Current quantization techniques for LLMs mainly fall into quantization-aware training (QAT) \cite{ke2024dl} and post-training quantization (PTQ) frameworks \cite{guan2024aptq,zhao2024lrquant,zhu2024survey}.
	
	\textbf{Quantization-Aware Training (QAT):} QAT involves simulating the quantization effect during the training process itself \cite{ke2024dl}. This is done to mitigate the performance degradation introduced by the quantization of parameters. Through QAT, the model can learn to adjust its weights to accommodate the lower precision, potentially preserving more accuracy (compared to PTQ) at the cost of substantially more intensive computations. QAT usually requires fine-tuning the model with quantization-aware layers that mimic the behavior of quantization during forward passes but use full precision in backward passes for gradient calculations.
	
	\textbf{Post-Training Quantization (PTQ):} In PTQ, quantization is directly applied to the existing pre-trained model without the need for retraining \cite{guan2024aptq,li2024norm}. Some PTQ methods require a small calibration dataset to calculate the activation magnitudes or identify \textit{salient weights} \cite{yao2024exploring}. Salient weights are those that have a large effect on the output of either the layer, block, or the final output of the model. PTQ-based methods are generally faster and more resource-efficient compared to QAT-based methods \cite{gong2024makes}. However, without retraining, these methods must rely on heuristics in the quantization process, such as the choice of the scaling factors and the design of the outlier retention \cite{guan2024aptq,li2024norm}. While they have lower computational costs, PTQ methods often result in larger performance degradation compared to QAT methods. Particularly, it is suggested that the performance of models that are quantized to fewer than 3 bits by most PTQ methods degrades substantially \cite{xu2024onebit}.
	
	Representative approaches for PTQ and QAT are reviewed in Appendix \ref{intro:recent_development_in_ptq_and_qat}, and techniques for alleviating QAT’s computational burden are detailed in Appendix \ref{s:efficient-qat}.
	
	\subsection*{Combining QAT with Parameter-Efficient Fine-Tuning}
	A recent line of work \cite{dettmers2024qlora,qin2024accurate,jeon2024l4q} combines QAT retraining with Parameter-Efficient Fine-Tuning (PEFT) methods, such as Low-Rank Adaptation (LoRA) \cite{kim2024ra}. PEFT can serve two purposes: (1) recovering the performance loss caused by quantization and (2) fine-tuning the quantized model for better adaptation to downstream tasks.
	
	QLoRA \cite{dettmers2024qlora} is the pioneering work in this direction. This approach integrates quantization and PEFT in a straightforward manner. The model is first quantized using a standard PTQ method. Then, the quantized LLM undergoes PEFT, where the quantized weights remain fixed, and only the LoRA parameters are fine-tuned to adapt to the specific task. Although QLoRA substantially reduces GPU memory and fine-tuning time while maintaining strong performance at the 4-bit level, it suffers from considerable performance degradation below 4 bits.
	

	ApiQ (Activation-preserved initialization of Quantized LLMs) \cite{liao2024apiq} aims to preserve the activations of the full-precision model by jointly optimizing the quantization parameters and the initial values of LoRA weights, rather than simply setting the initial values to zeros. This optimization is performed layer- or block-wise, proceeding sequentially from the lower layers to ensure that the output of each layer (or block) closely aligns with the original output. This prevents the accumulation of quantization errors across layers (or blocks). The layer- or block-wise training approach also reduces the memory requirements for fine-tuning. We provide more technical details on ApiQ in Section \ref{sec:apiq_method_description}. 

	\section{Technical background}
	
	\subsection{Toward ultra-low-bit quantization}
	\label{sec:intro_ultra_low_bit}
	Both QAT- and PTQ-based methods have succeeded in nearly preserving the performance of full-precision models at 4-bit precision \cite{wang2024fp4}. However, there remains substantial room for improvement as the bit-width is reduced to 3, 2, or 1 bit (ultra-low bit). 
	In the ultra-low-bit quantization regime, a major challenge is managing the salient weights, as they can unnecessarily extend the quantization range \cite{shang2023pb}. Several primary strategies are commonly employed to address this challenge. The first approach involves retaining salient weights in higher precision, which results in a mixed-precision model \cite{shang2023pb,huang2024billm}. The second approach is to increase the expressiveness of the quantization formulation to better accommodate salient weights \cite{chen2024db,xu2024onebit,jo2024mixture}. Here, some of the prominent approaches tackling the ultra-low-bit quantization problem are reviewed.
	
	PB-LLM \cite{shang2023pb} was the first work on 1-bit quantization for LLMs. It first filters out salient weights based on their magnitude (or their Hessian matrix) and then quantizes the remaining non-salient weights using standard PTQ methods such as GPTQ \cite{frantar2022gptq}. It also explores a QAT-based approach where the salient weights are frozen, and only the non-salient weights are trained. However, to achieve competitive performance, this method requires storing up to 30\% of all weights as salient weights, effectively increasing the bit-width and necessitating unstructured mixed-precision operations.
	
	DB-LLM \cite{chen2024db} explores the second perspective of increasing the expressiveness of the quantization formulation, specifically targeting the 2-bit regime. This method introduces a flexible dual binarization by representing 2-bit quantized weights as the scaled sum of two binary sets. By optimizing the scaling factors for each binary set, it enables flexible non-uniform quantization. During inference, this approach benefits from the efficiency of binary computations and the sparsity of binary representations.
	
	OneBit \cite{xu2024onebit} also addresses this issue by increasing the expressiveness of the quantization formulation. It incorporates scaling factors for both the input and output dimensions, thereby achieving a more flexible representation of weights than RTN. OneBit approach eliminates the need for storing separate information about the salient weights, reducing memory overhead and simplifying the inference process.
	
	Building on OneBit \cite{xu2024onebit}, BinaryMoS \cite{jo2024mixture} draws inspiration from a mixture-of-experts approach by introducing multiple scaling experts and a \textit{router}\footnote{A router is a linear layer that outputs a weight for each scaling expert based on the input token.} 
	that dynamically combines them based on the input token. The BinaryMoS method showed that adding multiple scaling factors only requires minimal memory overhead while substantially enhancing the representational capacity of the scaling factors \cite{jo2024mixture}.

	\subsection{The ApiQ method as a foundation}\label{sec:apiq_method_description}
	Results on the perplexity of the reviewed quantization methods are provided in Table \ref{tab:llama2_comparison} (in the Appendices) based on two LLMs: LLaMA-2-7B and LLaMA-2-13B. As shown in Table \ref{tab:llama2_comparison}, the LoRA weights used in ApiQ can effectively mitigate quantization errors in low-bit regimes (down to 3 bits and 2 bits). ApiQ has been shown to outperform strong PTQ baselines such as OmniQuant \cite{shao2023omniquant} and AWQ \cite{lin2024awq}, along with preceding quantization methods integrating PEFT, such as QLoRA \cite{dettmers2024qlora} and LoftQ \cite{li2023loftq}. The ApiQ method is therefore chosen as a foundation for our proposed method. ApiQ comprises two stages: (1) a post-training quantization stage, and (2) a fine-tuning stage.
	
	ApiQ was motivated by the observation that under PEFT, preserving the starting point and mitigating the propagation of quantization error from shallower layers to deeper layers are crucial for achieving better accuracy \cite{liao2023make}. Given that LoRA \cite{hu2022lora} is arguably one of the most widely used PEFT methods , we will limit the scope of our discussion to LoRA. In traditional PEFT without quantization, LoRA weights initialized to zeros inherently preserve the initial state. However, through quantization, the original weights undergo substantial modifications, particularly in ultra-low-bit settings. This requires finding a more effective initialization for LoRA weights to ensure that the combined weights remain as close as possible to the original full-precision weights.
	
	To achieve the aforementioned objectives while conserving memory and computational resources during training, ApiQ employs block-wise training. In this approach, for each block (e.g., each transformer block), the quantization parameters (such as weight clippings) and LoRA weights of all layers within the block are jointly optimized to minimize the output error, as defined in Equation \ref{eq:ApiQ_bw_objective}.
	\begin{equation}
		\label{eq:ApiQ_bw_objective}
		\arg\min_{Q_s, A_s, B_s} \left\| \mathcal{F}(\mathbf{W}_s, \mathbf{X}) - \mathcal{F}(\mathbf{Q}_s, \mathbf{A}_s, \mathbf{B}_s, \mathbf{X}^q) \right\|
	\end{equation}
	
	In Equation \ref{eq:ApiQ_bw_objective}, $\mathcal{F}$ denotes the mapping function of a transformer block, 
	$\mathbf{W}_s$ represents all the weights of the linear layers within this block, 
	$\mathbf{X}$ is the input to this block, 
	$\mathbf{Q}_s$ are the quantized versions of $\mathbf{W}_s$, 
	$\mathbf{A}_s$ and $\mathbf{B}_s$ are all low-rank matrices within this block, 
	and $\mathbf{X}^q$ is the input to the quantized block, which is also the output from the preceding quantized block.
	
	ApiQ is also compared to previous methods such as LoftQ \cite{li2023loftq} in Table \ref{tab:llama2_comparison} of Appendices. LoftQ, which optimizes quantization parameters for each layer independently with the objective of preserving the original weights. Compared to LoftQ, ApiQ offers the advantage of mitigating the propagation of quantization errors. This is achieved by allowing errors introduced in one layer to be further minimized through optimization in the subsequent layer, at the expense of requiring additional time for training.
	
	In our usage of ApiQ as a foundation, we will focus on the first stage of ApiQ, viewing it as a PTQ method. Specifically, we optimize only the initialized PEFT parameters (i.e. LoRA weights) via the block-wise training procedure described, without conducting additional fine-tuning on downstream tasks.
	
	\section{Investigation of preliminary ideas}\label{sec:methods_explored}

	As mentioned in Section \ref{sec:apiq_method_description}, ApiQ has demonstrated strong performance under 2-bit quantization through its novel approach. However, ApiQ employs a relatively simplistic quantization framework, where the clipping ranges are the sole learnable parameters. This limitation suggests the potential for performance improvement by adopting more expressive quantization formulations including those proposed in other existing works \cite{jo2024mixture,chen2024db}. 
	
	It is noteworthy that several high-performing methods for ultra-low-bit quantization rely on quantization-aware training (QAT) \cite{jo2024mixture,chen2024db}, which entails substantial computational and memory overhead. By leveraging the strong representational capacity of quantization algorithms from these methods and combining it with the low training cost of ApiQ, it may be possible to strike a balance between performance and computational cost in a new quantization method.
	
	To investigate this hypothesis, we incorporated quantization formulations from two recently proposed and high-performing methods: BinaryMoS \cite{jo2024mixture} and DB-LLM \cite{chen2024db}. Although these investigations ultimately proved ineffective as shown in Appendix \ref{sec:result_preliminary_ideas}, they provided us important insights and paved the way for our proposed method (discussed in Section \ref{sec:proposed_methods}). 
	
	\subsection{Combining ApiQ with BinaryMoS}\label{sec:integration_of_binarymos}
	BinaryMoS \cite{jo2024mixture} has outperformed all its competitors in terms of accuracy for 1-bit quantization. It is inspired by the \textit{mixture of experts} approach, incorporating multiple scaling experts and a router that dynamically combines them based on the input token.
	
	However, as a QAT-based approach, BinaryMoS incurs relatively high computational and memory costs. Therefore, we hypothesize that leveraging LoRA weights to compensate for accuracy loss, instead of retraining the original weights, could potentially reduce training time and/or memory consumption.
	
	To investigate this idea, we replaced the quantized linear layer in the ApiQ codebase with its counterpart from BinaryMoS and trained the quantization parameters alongside LoRA weights in a block-wise manner, following the same approach as ApiQ (described in Section \ref{sec:apiq_method_description}). The results for this investigation are provided in Appendix \ref{sec:ApiQ+BinaryMoS_results}. Contrary to our initial hypothesis, our combined method exhibited substantially poorer performance than the original BinaryMoS approach. This result suggests an inherent limitation in the ability of LoRA weights to counteract the accuracy loss induced by quantization, especially under the stringent constraint of 1-bit quantization.

	\subsection{Combining ApiQ with DB-LLM}\label{sec:integration_of_db_llm}
	DB-LLM \cite{chen2024db} is a method specifically designed for 2-bit quantization and has some of the best 2-bit performance measures among the techniques reviewed in Section \ref{sec:intro}. This method introduces flexible dual binarization, which represents 2-bit quantized weights as a scaled sum of two binary sets. By optimizing the scaling factors for each binary set, it enables flexible and non-uniform quantization.

	Following the same approach that we had with BinaryMoS, we replaced the quantized linear layer in ApiQ with its counterpart from DB-LLM and applied the same training procedure as used in ApiQ. The results for this approach are provided in Appendix \ref{sec:preliminary_exp_db_llm}. We found that applying the DB-LLM quantization formulation augments the model’s expressive power. However, it also heightens its susceptibility to overfitting, causing the quantized model to perform worse when trained on a dataset of the same size.
	


	\section{Proposed method}\label{sec:proposed_methods}
	Like most methods that rely on a calibration dataset, ApiQ is prone to overfitting to the calibration dataset. A sign of overfitting can be observed in the perplexity gap between the WikiText-2 and C4 datasets, which is shown in Fig.\ \ref{fig:reproduced_results} of Appendices. For the model quantized with ApiQ, this gap becomes larger than that of the original full-precision model. We focused on addressing this issue while preserving its advantage of mitigating output errors. 
	
	
	A widely used approach to addressing the overfitting problem is the addition of a regularization term. Since this method does not require additional data or introduce new trainable parameters, it should incur minimal memory and computational overhead, making it a suitable component for our proposed approach. 
	
	\subsection{Preservation of original weights}\label{ss:reg2} We first investigated a naive regularization approach that involves preserving the original weights by incorporating the L2 distance between the original and quantized weights into the loss function. EasyQuant \cite{tang2024easyquant} adopted this regularization term as the only term in loss function and demonstrated promising results for 4-bit quantization. Notably, this approach eliminates the risk of overfitting, as it does not depend on values from the calibration dataset. Moreover, it is intuitively reasonable that preserving the original weights as much as possible contributes to maintaining the original performance. We evaluated two variants of this regularization strategy. 
	
	In the first variant, the penalty term is defined as the difference between the original weights and the quantized weights before adding the LoRA component (hereafter referred to as "BeforeLoRA"). In the second variant, the penalty term is defined as the difference between the original weights and the final weights after the LoRA adjustments have been applied (hereafter referred to as "AfterLoRA"). The results are provided in Appendix \ref{sec:result_reg_original_weights_preservation}. In short, even with hyperparameter tuning, this naive regularization does not exceed the performance of the original ApiQ method, motivating our saliency-aware approach.
	
	\subsection{Saliency-aware preservation of original weights}
	\label{ss:reg3}
	The previous naive weight preservation method assigns equal importance to all parameters within a weight matrix (or a subset thereof). However, as demonstrated in prior studies such as SqueezeLLM \cite{kim2023squeezellm}, certain parameters (the salient parameters) have a greater impact on the model's output. At this stage, we hypothesize that prioritizing the preservation of these salient weights leads to improved model performance. To investigate this idea, we use the regularization term in Eq.\ \eqref{eq:reg}, where $w_{i}$ refers to each parameter in the weight matrix, $Q$ denotes the quantization function, and $\alpha_{i}$ represents the saliency of the parameter, which is used as its weight in the regularization term.
	\begin{equation}
		\label{eq:reg}
		\sum_{i=1}^{N} \alpha_i (w_i - Q(w_i))^2.
	\end{equation}
	We calculated $\alpha_{i}$ in the same manner as SqueezeLLM \cite{kim2023squeezellm}, which approximates the Hessian matrix using squared gradients. The gradients were calculated using a small calibration set of 100 samples, each with a sequence length of 512, drawn from either the WikiText-2 or C4 dataset.
	The results are shown in Section \ref{sec:result_reg_saliency_aware_weights_preservation}.

	\subsection{Training framework}
\begin{figure}[ht]
		\centering
		\begin{subfigure}[b]{0.36\linewidth}
			\centering
			\includegraphics[width=\linewidth]{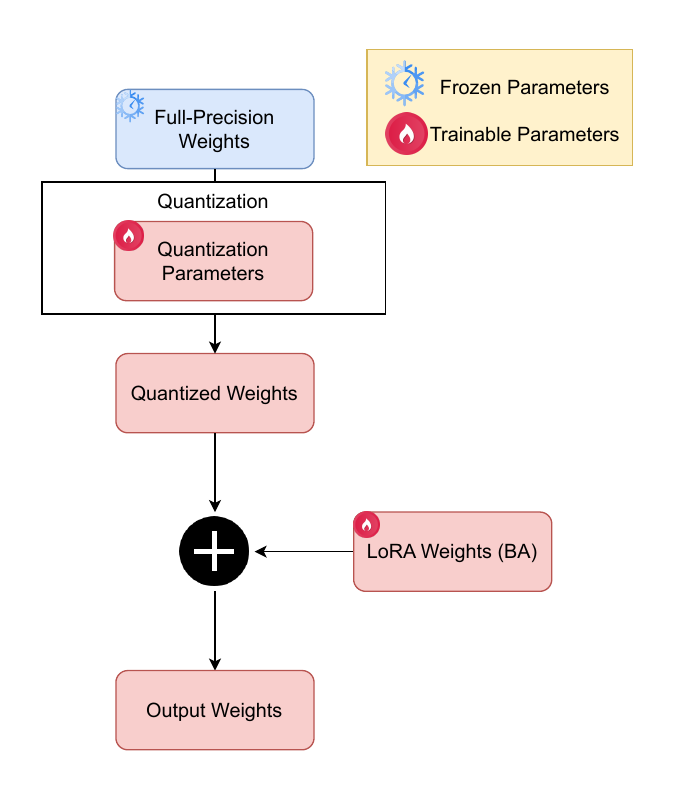}
			\caption{The process for obtaining output weights}
			\label{fig:weight_computation}
		\end{subfigure}
		\hfill
		\begin{subfigure}[b]{0.63\linewidth}
			\centering
			\includegraphics[width=\linewidth]{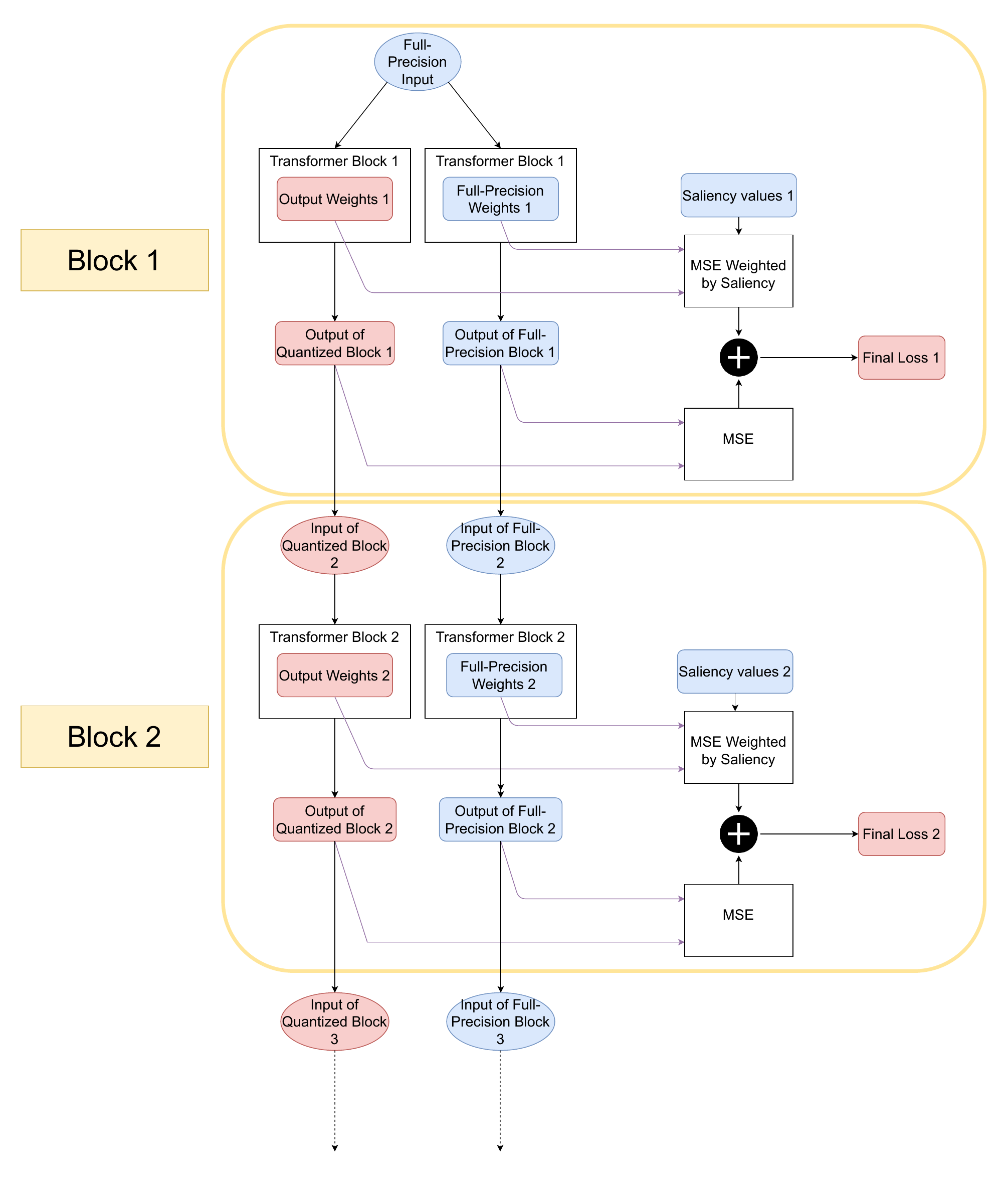}
			\caption{Block-wise training flow}
			\label{fig:block_wise_training}
		\end{subfigure}
		\caption{Schematic overviews of the training framework. These high-resolution figures can be magnified on screen for the details.}
		\label{fig:flowcharts}
	\end{figure}

	The training framework is largely consistent with the approach described in Section \ref{sec:apiq_method_description}, with the only difference being the addition of a regularization term to the loss function. Schematic overviews of the training framework are presented in Figs.\ \ref{fig:weight_computation}--\ref{fig:block_wise_training}. Fig.\ \ref{fig:weight_computation} illustrates the process of obtaining the output weights (i.e., the weights used in the quantized model). In Figs. \ref{fig:weight_computation}--\ref{fig:block_wise_training}, rectangles with a red background indicate values that are updated during training, while those with a light blue background indicate values that remain fixed. The same color code is used in Fig.\ \ref{fig:block_wise_training} that depicts the flow of block-wise training. This procedure is applied sequentially from the first block to the final block (the 32nd block in the case of the LLaMA-2-7B model), using the output of the preceding block as the input for the next.
	
	

	\section{Experimental setup}\label{sec:experimental_setups}

	\subsection{Implementation details} 
	
	The ApiQ method requires a calibration dataset to optimize both the quantization parameters and the LoRA adapter weights. For this purpose, we randomly sampled a subset of sentences from the training set of WikiText-2 \cite{merity2017pointer}. Given its superior accuracy and time efficiency compared to layer-wise optimization\cite{liao2024apiq}, we conducted our experiments exclusively using block-wise optimization. In all experiments, we used the hyperparameter configuration listed in Table \ref{tab:hyperparameters} of Appendices, unless stated otherwise.

	\subsection{Target language models} The majority of our evaluations were conducted on the LLaMA-2-7B model. However, to confirm the consistency of our proposed method, additional experiments were performed on the LLaMA-2-13B model, as presented in part (b) of Table \ref{tab:ApiQ_saliency_aware_weight_preservation_reg}. Our LLaMA-2-13B experiments were mainly centred on evaluating the saliency-aware weight preservation term, which—as shown in Section \ref{sec:result_reg_saliency_aware_weights_preservation}—demonstrated promising results on the 7B model.

	\subsection{Metrics and datasets used}

	To assess the effectiveness of quantization, namely the extent to which the accuracy of the original full-precision model is preserved, we used perplexity (PPL) as the evaluation metric. To align with previous research, we used the test set of WikiText-2 \cite{merity2017pointer} and validation set of C4 \cite{raffel2020exploring}. Also, consistent with the literature, we assessed the zero-shot accuracy on five common-sense reasoning tasks, namely WinoGrande \cite{sakaguchi2021winogrande}, PIQA \cite{bisk2020piqa}, HellaSwag \cite{zellers2019hellaswag}, Arc-Easy \cite{clark2018think}, and Arc-Challenge \cite{clark2018think}.

	\section{Results}

	\subsection{Evaluation of the proposed method}
	\label{sec:result_reg_saliency_aware_weights_preservation}
	
	In this section, we provide the results of the saliency-aware weight preservation term proposed in Section \ref{ss:reg3}. The results of the naive weight preservation term (introduced in Section \ref{ss:reg2}) are provided in Appendix \ref{sec:result_reg_original_weights_preservation}.

	\begin{table}
		\centering
		\caption{Performance of ApiQ with saliency-aware weight preservation regularization on LLaMA-2 models}
		\label{tab:ApiQ_saliency_aware_weight_preservation_reg}
		
		\begin{subtable}[t]{\textwidth}
			\centering
			\caption{7B model}
			\resizebox{\textwidth}{!}{%
				\begin{tabular}{c c c c c c c c c c c}
					\toprule
					\multirow{2}{*}{\textbf{LoRA Variant|}} 
					&
					\multirow{2}{*}{\textbf{Calibration Dataset|}}
					& \multirow{2}{*}{\textbf{Coefficient}}
					& \multicolumn{2}{r}{\textbf{PPL}\,$\downarrow$}
					& \multicolumn{6}{c}{\textbf{Accuracy (\%)}\,$\uparrow$} \\
					\cmidrule(lr){4-5} 
					\cmidrule(lr){6-11}
					& & & \textbf{WikiText2} & \textbf{C4}
					& \textbf{WinoGrande} & \textbf{HellaSwag} 
					& \textbf{ArcC}    & \textbf{ArcE}
					& \textbf{PiQA}    & \textbf{Average} \\
					\midrule
					\multicolumn{3}{c}{\textbf{Full-Precision Model}} & 5.47 & 6.97 & 69.22 & 57.16 & 43.52 & 76.26 & 78.07 & 64.85 \\
					\multicolumn{3}{c}{\textbf{Original ApiQ (No regularization)}} & 7.56 & 10.42 & 62.83 & 46.53 & 31.57 & 66.54 & 72.09 & 55.91 \\
					\midrule
					AfterLoRA & WikiText-2 & 1 & 7.54 & \textbf{10.18} & \textbf{64.09} & \textbf{47.08} & 32.08 & \textbf{67.55} & \textbf{72.63} & \textbf{56.69} \\
					AfterLoRA & C4 & 1 & \textbf{7.53} & 10.34 & 63.22 & 46.80 & 32.42 & 66.67 & 72.25 & 56.27 \\
					\midrule
					BeforeLoRA & WikiText-2 & 1 & 7.58 & 10.65 & 63.46 & 46.35 & 32.25 & 67.13 & 71.71 & 56.18 \\
					BeforeLoRA & C4 & 1 & 7.57 & 10.38 & 62.59 & 46.12 & \textbf{32.85} & 67.17 & 72.20 & 56.18 \\
					\bottomrule
				\end{tabular}
			}
		\end{subtable}
		
		\begin{subtable}[t]{\textwidth}
			\centering
			\caption{13B model}
			\resizebox{\textwidth}{!}{%
				\begin{tabular}{c c c c c c c c c c c}
					\toprule
					\multirow{2}{*}{\textbf{LoRA Variant|}} 
					\multirow{2}{*}{\textbf{Calibration Dataset|}}
					& \multirow{2}{*}{\textbf{Coefficient}}
					& \multicolumn{2}{r}{\textbf{PPL}\,$\downarrow$}
					& \multicolumn{6}{c}{\textbf{Accuracy (\%)}\,$\uparrow$} \\
					\cmidrule(lr){4-5} 
					\cmidrule(lr){6-11}
					& & & \textbf{WikiText2} & \textbf{C4}
					& \textbf{WinoGrande} & \textbf{HellaSwag} 
					& \textbf{ArcC}    & \textbf{ArcE}
					& \textbf{PiQA}    & \textbf{Average} \\
					\midrule
					\multicolumn{3}{c}{\textbf{Full-Precision Model}} & 4.88 & 6.47 & 72.22 & 60.07 & 48.29 & 79.42 & 79.05 & 67.81 \\
					\multicolumn{3}{c}{\textbf{Original ApiQ (No regularization)}} & 6.46 & 8.93 & 67.72 & 51.70 & 37.29 & 73.02 & 74.81 & 60.91 \\
					\midrule
					AfterLoRA & WikiText-2 & 2 & 6.45 & \textbf{8.90} & 67.64 & 51.98 & 38.23 & \textbf{73.48} & \textbf{75.73} & 61.41 \\
					AfterLoRA & C4 & 2 & \textbf{6.43} & 9.66 & \textbf{67.96} & \textbf{52.01} & \textbf{38.65} & 73.06 & 75.46 & \textbf{61.43} \\
					\bottomrule
				\end{tabular}
			}
		\end{subtable}
		
	\end{table}

	We tested using different coefficients for the regularization term and evaluated the two variants (Before LoRA and After LoRA described in Section \ref{ss:reg2}). Additionally, we experimented with different calibration datasets (C4 and WikiText-2). The results are provided in Tables \ref{tab:ApiQ_saliency_aware_weight_preservation_reg}--\ref{tab:ApiQ_saliency_aware_weight_preservation_reg_coeffs}. Table \ref{tab:ApiQ_saliency_aware_weight_preservation_reg} presents the comparison between the two variants and different calibration datasets. Table \ref{tab:ApiQ_saliency_aware_weight_preservation_reg_coeffs} shows the results when varying the regularization term coefficient, with the After-LoRA variant and WikiText-2 calibration dataset fixed.

	\begin{table}[ht]
		\centering
		\caption{Performance of ApiQ with saliency-aware weight preservation regularization on LlaMA-2-7B model (with different coefficients for the regularization term)}
		\label{tab:ApiQ_saliency_aware_weight_preservation_reg_coeffs}
		\resizebox{\textwidth}{!}{%
			\begin{tabular}{cc|cc|cccccc}
				\toprule
				& & \multicolumn{2}{c|}{\textbf{PPL}\,$\downarrow$} & \multicolumn{6}{c}{\textbf{Accuracy (\%)}\,$\uparrow$} \\
				\cmidrule(lr){3-4}\cmidrule(lr){5-10}
				\textbf{Initial Coefficient|} & \textbf{Coefficient Multiplier} & \textbf{WikiText2} & \textbf{C4} & \textbf{WinoGrande} & \textbf{HellaSwag} & \textbf{ArcC} & \textbf{ArcE} & \textbf{PiQA} & \textbf{Average} \\
				\midrule
				\multicolumn{2}{c|}{\textbf{Full-Precision Model}} & 5.47 & 6.97 & 69.22 & 57.16 & 43.52 & 76.26 & 78.07 & 64.85 \\
				\multicolumn{2}{c|}{\textbf{Original ApiQ (No regularization)}} & 7.56 & 10.42 & 62.83 & 46.53 & 31.57 & 66.54 & 72.09 & 55.91 \\
				\midrule
				1.0 & 1.0 & 7.54 & \textbf{10.18} & 64.09 & 47.08 & 32.08 & \textbf{67.55} & 72.63 & 56.69 \\
				2.0 & 1.0 & 7.57 & 10.26 & \textbf{64.80} & \textbf{47.11} & 32.34 & 67.21 & \textbf{72.96} & \textbf{56.88} \\
				10.0 & 1.0 & 7.72 & 10.75 & 63.38 & 44.99 & 31.31 & 66.04 & 71.24 & 55.40 \\
				\midrule
				1e-3 & 1.3 & \textbf{7.53} & 10.41 & 64.25 & 46.82 & \textbf{33.02} & 66.04 & 72.63 & 56.55 \\
				1e-2 & 1.3 & 7.54 & 10.36 & 62.98 & 46.90 & 32.00 & 65.61 & 72.52 & 56.00 \\
				1e-1 & 1.3 & 7.62 & 10.43 & 62.51 & 46.03 & 32.68 & 67.51 & 72.58 & 56.26 \\
				\bottomrule
			\end{tabular}
		}
	\end{table}

	In Table \ref{tab:ApiQ_saliency_aware_weight_preservation_reg_coeffs}, the coefficient multiplier column indicates the factor by which the regularization term coefficient is scaled for each layer\footnote{This adjustment is intended to account for the increasing quantization loss in deeper layers, ensuring that the regularization term is aligned with the loss.}.
	
	Across a wide range of hyperparameters, the inclusion of this regularization term enhances the average zero-shot accuracy of the quantized model by up to 0.97 \% in absolute terms for LLaMA-2-7B model as shown in Tables \ref{tab:ApiQ_saliency_aware_weight_preservation_reg}--\ref{tab:ApiQ_saliency_aware_weight_preservation_reg_coeffs}. This improvement covers 10.85\% of the accuracy gap between the ApiQ method and the full-precision model of LLaMA-2-7B. 
	
	Similarly, for LLaMA-2-13B model, we observed consistent improvements across all few-shot tasks, with an average zero-shot accuracy increase of 0.52\% in absolute terms. This corresponds to 7.54\% of the accuracy gap, further confirming the consistency and effectiveness of our approach.

	
	\section{Discussions}
	\label{ch:discussion}
	We pursued two primary lines of investigation: (1) examining whether existing QAT-based quantization methods (specifically BinaryMoS \cite{jo2024mixture} and DB-LLM \cite{chen2024db}) could improve an ApiQ-based quantization method \cite{liao2024apiq} under constrained retraining; and (2) exploring saliency-aware weight preservation regularization to enhance ApiQ’s accuracy by mitigating overfitting. The results and implications from the first line of investigation in Appendix \ref{sec:result_preliminary_ideas}.
	In summary, we confirmed that maintaining high accuracy in QAT-based methods requires retraining the original model weights to ensure sufficient representational capacity, and leveraging large, diverse datasets to prevent overfitting.

	Our second investigation was centered on the trade-off between two common training objectives—weight preservation and output preservation—within the framework of ApiQ. Despite the effectiveness of naive weight preservation under 4-bit quantization \cite{li2023loftq}, ApiQ revealed that this approach performs poorly under 2-bit quantization, with perplexity of quantized LLaMA-2-7B model exceeding 500 on WikiText-2 and C4 datasets. This can be attributed to the accelerated accumulation of output errors in deeper layers. While ApiQ achieved substantially better performance in the 2-bit setting, it exhibited noticeable overfitting. We argue that the overfitting of ApiQ can be attributed to its exclusive reliance on output preservation, which is highly dependent on the quality and size of the calibration data. To address this drawback of ApiQ, our proposed method incorporates both weight and output preservation objectives.
	
	The results in Appendix \ref{sec:result_reg_original_weights_preservation} show that introducing a naive weight preservation objective leads to degraded performance. This degradation in performance is consistent with the observed ineffectiveness of the LoftQ method under 2-bit quantization \cite{li2023loftq}.
	
	On the other hand, experimental results in Section \ref{sec:result_reg_saliency_aware_weights_preservation} showed that incorporating our proposed saliency-aware regularization term yielded modest yet consistent improvements in perplexity and zero-shot accuracy— recapturing 7.54\% and 10.85\% of the performance gap between baseline ApiQ and the original model of LLaMA-2-7B and LLaMA-2-13B respectively. This performance boost verifies that using the preservation of salient weights as a training target can be effective in reducing the accumulation of output errors and alleviating overfitting to the training data at the same time.
	
	The improvement achieved by saliency-aware weight preservation over naive weight preservation is aligned with the insights from the SqueezeLLM model \cite{kim2023squeezellm}. In particular, our results reaffirm the results from \cite{kim2023squeezellm} highlighting the benefits of assigning greater importance to salient weights in non-uniform quantization, particularly under 4- bit and 3-bit compression regimes.
	
	Furthermore, the overhead in memory was negligible in our proposed method tested on the quantization of LLaMA-2-7B and LLaMA-2-13B models.
	Unlike the memory overhead, runtime was observed to increase in our proposed method due to additional gradient computations required. Taken together, our findings indicate that focusing on salient weights helps preserve crucial representational capacity under severe bit-width constraints, strengthening the potential of partial retraining strategies such as ApiQ.

	\section{Conclusions}
	\label{ch:conclusion}
	
	In this study, we conducted two lines of experiments focusing on design considerations for improving an existing frontier quantization method in the ultra-low-bit regimes.
	
	As discussed in Appendix \ref{sec:discussion_QAT_plus_ApiQ}, through our first set of experiments, we were able to gain insight into the underlying factors contributing to the effectiveness of QAT methods. First, by integrating BinaryMoS into ApiQ, we observed that under 1-bit quantization, retraining the model weights offers a representational capacity that cannot be achieved by training LoRA adapters alone. This retraining — despite being memory and compute intensive — is critical for preserving model performance. Second, by integrating DB-LLM into ApiQ, we reaffirmed findings from EfficientQAT \cite{chen2024efficientqat} that highlight the importance of using large and diverse datasets to mitigate overfitting and maintain accuracy.
	
	In our second line of investigation, we examined the trade-off between output preservation and weight preservation—two major training objectives used in both PTQ and QAT. Under 2-bit quantization, we found that incorporating a naive weight preservation regularization term had a negative effect on the accuracy of the quantized model. Informed by this finding, our proposed saliency-aware weight preservation term assigns greater importance to weights most influential on the model’s output. Applying our proposed method on several LLMs, we observed consistent improvement in zero-shot accuracies across tasks.

	Although we demonstrated the practicality of our proposed saliency-aware weight preservation regularization term in the specific context of ApiQ, our approach is generalizable to any quantization method that involves some form of training such as LLM-QAT \cite{liu2023llm} and OneBit \cite{xu2024onebit}.
	
	One potential direction for future work is to explore different orders of applying the two training objectives (output preservation and saliency-aware weight preservation), and more flexible combinations of them. For example, one could first obtain a suitable initialization by training with the saliency-aware weight preservation objective, and then fine-tune the quantization parameters using output preservation as the training objective. When combined with the two training paradigms (block-wise training and end-to-end training as outlined in EfficientQAT \cite{chen2024efficientqat}), these objectives are expected to offer additional flexibility for exploration. For instance, the saliency-aware weight preservation term can be applied exclusively during block-wise training, exclusively during end-to-end training, or during both stages. Further research is needed to determine which of these approaches will tighten the accuracy gap more effectively.
	
	Another avenue for future work is to explore alternative definitions of saliency.
	In our study, saliency was defined as the degree to which each weight influences the model’s final output. However, an alternative definition could consider the impact of each weight on the output of its corresponding layer or block \cite{lin2024awq}. Future research can investigate if alternative definitions of saliency enable a more fine-grained control during optimization and become more effective for a desired outcome.

	\FloatBarrier

	\section*{Acknowledgments}
	The authors thank the three anonymous reviewers of the MDAI'25 for their valuable comments and Baohao Liao for helpful insights on the ApiQ method.

	

	
	\appendix
	\section*{Appendices}

	\section{Experimental details}
	In Table \ref{tab:hyperparameters}, we present the complete hyperparameter configurations used throughout our experiments unless otherwise noted.
	\begin{table}
		\centering
		\caption{Hyperparameter configuration}
		\begin{tabular}{ll}
			\toprule
			\textbf{Hyperparameter}           & \textbf{Value} \\ 
			\midrule
			\multicolumn{2}{l}{\textbf{Granularity and Bit-width}} \\ 
			Optimization granularity          & block-wise   \\ 
			Bit-width                  & 2       \\ 
			\midrule
			\multicolumn{2}{l}{\textbf{LoRA Settings}} \\ 
			LoRA rank        & 64     \\ 
			\midrule
			\multicolumn{2}{l}{\textbf{Optimizer and Training Settings}} \\ 
			Optimizer                  & AdamW     \\ 
			Weight decay for quantization parameters  & 0.1      \\ 
			Learning rate for quantization parameters  & 0.005     \\ 
			Weight decay for LoRA weights        & 0.1      \\ 
			Learning rate for LoRA weights       & 0.0005     \\ 
			Batch size                 & 1       \\ 
			Epochs (per layer)                  & 20       \\ 
			\midrule
			\multicolumn{2}{l}{\textbf{Calibration Dataset Settings}} \\ 
			Sequence length for calibration       & 2048      \\ 
			Number of calibration samples        & 128      \\ 
			\bottomrule
		\end{tabular}
		\label{tab:hyperparameters}
	\end{table}
	
	\section{Computing resources}
	The majority of experiments were conducted on a single NVIDIA A100 GPU with 40GB of memory. However, for computing the squared gradients required by the LLaMA-2-13B model, we employed a single NVIDIA H100 GPU with 80GB of memory due to the substantial memory demands of this computation.
	
	\begin{table}[ht]
		\centering
		\caption{The duration and peak GPU memory used for quantizing LLaMA-2 models}
		\label{tab:resource_usage}
		
		\begin{subtable}[t]{0.9\textwidth}
			\centering
			\caption{7B}
			\rowcolors{2}{gray!15}{white}
			\begin{tabular}{
					>{\raggedright\arraybackslash}m{6.5cm}
					>{\centering\arraybackslash}m{2cm}
					>{\centering\arraybackslash}m{2cm}
				}
				\toprule
				\textbf{Method} & \textbf{Duration} & \textbf{Peak GPU memory} \\
				\midrule
				ApiQ (Original) & 1.5h & 16GB \\
				ApiQ With Saliency-Aware Weight Preservation Regularization (After LoRA) & 2.0h & 16GB \\
				ApiQ With Saliency-Aware Weight Preservation Regularization (Before LoRA) & 1.8h & 16GB \\
				ApiQ + DB-LLM & 1.6h & 16GB \\
				ApiQ + DB-LLM With Saliency-Aware Weight Preservation Regularization (After LoRA) & 2.2h & 16GB \\
				\bottomrule
			\end{tabular}
		\end{subtable}
		
		\vspace{1em}
		
		\begin{subtable}[t]{0.9\textwidth}
			\centering
			\caption{13B}
			\rowcolors{2}{gray!15}{white}
			\begin{tabular}{
					>{\raggedright\arraybackslash}m{6.5cm}
					>{\centering\arraybackslash}m{2cm}
					>{\centering\arraybackslash}m{2cm}
				}
				\toprule
				\textbf{Method} & \textbf{Duration} & \textbf{Peak GPU memory} \\
				\midrule
				ApiQ (Original) & 2.7h & 29GB \\
				ApiQ With Saliency-Aware Weight Preservation Regularization (After LoRA) & 3.8h & 29GB \\
				\bottomrule
			\end{tabular}
		\end{subtable}
		
	\end{table}
	
	The memory usage and training time of some of the experiments are provided in Table \ref{tab:resource_usage}. The proposed regularization methods introduce negligible GPU memory overhead; however, the training time increases by approximately 40\% due to the computation of the regularization terms. This could be reduced through further optimization. It is important to note that quantization is usually performed only once, whereas inference is carried out repeatedly in deployment. Consequently, the modest extra time required for quantization is quickly amortized over a limited number of inference runs, and the resulting boost in inference accuracy generally far outweighs this one-off overhead.
	
	Our work modifies only ApiQ’s training procedure and leaves inference unchanged. Therefore, we omit inference time and memory measurements. One may refer to the original ApiQ paper \cite{liao2024apiq} for their details on inference time and memory measurements.

	\FloatBarrier
	
	\section{Overview of round-to-nearest (RTN) quantization}
	\label{sec:rtn_example}
	Here, we illustrate the concept of quantization using the simplest approach—Round-To-Nearest (RTN) quantization. 
	Equation \ref{eq:quantization} describes the first part of quantization, which maps each original weight to an integer value between 0 and $2^{N-1}$, where N is the bit width of the quantized value.
	\begin{equation}
		\label{eq:quantization}
		\mathbf{W}_{\text{int}} = \text{clamp}\left( \left\lfloor \frac{\mathbf{W}}{s} \right\rceil + z,\ 0,\ 2^N - 1 \right)
	\end{equation}
	Here, $\lfloor \cdot \rceil$ represents the rounding operation. $\mathbf{W}_{\text{int}}$ and $\mathbf{W}$ denote the quantized integer and full-precision weight matrices, respectively. $s$ is the scaling factor and $z$ is the zero point. In the simplest case, these are computed using Equations \ref{eq:scaling_factor} and \ref{eq:zero_point}, where $W_{\max}$ and $W_{\min}$ represent the maximum and minimum values in the weight matrix, respectively.
	\begin{equation}
		\label{eq:scaling_factor}
		s = \frac{W_{\max} - W_{\min}}{2^N - 1}
	\end{equation}
	\begin{equation}
		\label{eq:zero_point}
		z = - \text{round}\left(\frac{W_{min}}{s}\right)
	\end{equation}
	
	In quantization, the scaling factor and zero point are typically shared among a group of weights—such as an entire weight matrix, a single channel within a matrix, or a subset of a channel. This grouping determines the granularity of quantization.
	
	During forward propagation, the quantized weights are converted back to full-precision values by applying the corresponding scaling factors and zero points—a process known as dequantization, as shown in Equation \ref{eq:dequantization}.
	\begin{equation}
		\label{eq:dequantization}
		\hat{\mathbf{W}} = (\mathbf{W}_{\text{int}} - z) \cdot s
	\end{equation}

	\begin{figure}[ht]
		\centering
		\includegraphics[width=\linewidth]{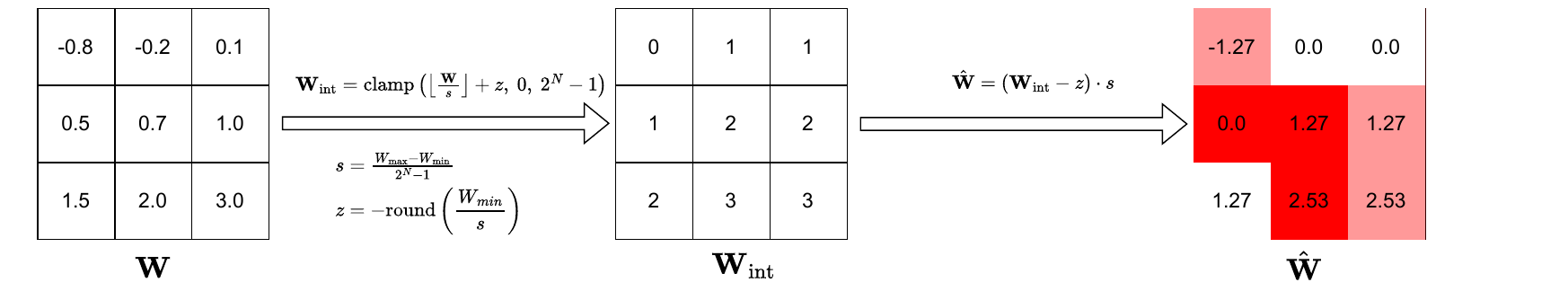}
		\caption{Numerical example of 2-bit quantization. The color intensity in the rightmost matrix corresponds to the deviation of the quantized weight from the original weight.}
		\label{fig:quantization_example}
	\end{figure}
	
	In Eq. \ref{eq:dequantization}, $\hat{\mathbf{W}}$ refers to the reconstructed weights used in the forward computation. Since our focus is on the reconstructed weights and how their deviation from the original weights affects model accuracy, we refer to the entire process of obtaining reconstructed weights $\hat{\mathbf{W}}$ from the original weights $\mathbf{W}$ as quantization. The whole process of quantization is illustrated in Fig.\ \ref{fig:quantization_example} using a numerical example.

	\FloatBarrier
	
	\section{A review of other related works}
	\subsection{Recent developments in PTQ and QAT}
	\label{intro:recent_development_in_ptq_and_qat}
	This section provides a review of recent developments in the area of PTQ \cite{kim2023squeezellm,lin2024awq} followed by notable works from the QAT category \cite{liu2023llm,du2024bitdistiller,xu2024onebit}.
	
	SqueezeLLM \cite{kim2023squeezellm} achieves near-lossless compression down to 3 bits by weighting parameters according to their impact on model output (sensitivity-weighted non-uniform quantization) and isolating outlier weights in a sparse matrix (dense and sparse decomposition). This combination preserves the representation of highly sensitive and outlier weights, effectively maintaining accuracy under aggressive quantization.
	
	AWQ \cite{lin2024awq} builds on the observation that retaining outlier weights in full precision can significantly improve performance. However, to avoid the hardware inefficiencies associated with mixed-precision formats, AWQ first identifies the output channels in each weight matrix with the largest activation magnitudes—referred to as salient channels—and then amplifies them. This approach leverages on quantization loss reduction through scaling up specific weights which was proven effective in comparisons provided against benchmarks \cite{lin2024awq}.
	
	LLM-QAT \cite{liu2023llm} is argued to be the first to apply a QAT framework to LLMs. Given the challenge of obtaining suitable fine-tuning datasets for retraining, LLM-QAT leverages generated data from the full-precision LLM itself and trains the quantized LLM to align with the output distribution of the full-precision model. This process is referred to as a knowledge distillation framework.

	BitDistiller \cite{du2024bitdistiller} adopts the same knowledge distillation framework as LLM-QAT \cite{liu2023llm} but enhances the naive symmetric quantization approach by using asymmetric quantization and customized clipping technique to preserve the fidelity of quantized weights. Specifically, before the main training, they use a small calibration set to optimize the asymmetric clipping thresholds for weights to minimize the output discrepancy caused by quantization. Additionally, it proposes a new self-distillation loss function\footnote{The loss function is named Confidence-Aware Kullback-Leibler Divergence \cite{du2024bitdistiller}.}, which takes into account the confidence of the full-precision model about each prediction. With these advancements, BitDistiller was shown to achieve substantial performance gains over existing methods in 3-bit and 2-bit quantization \cite{du2024bitdistiller}.
	
	OneBit \cite{xu2024onebit} was the first to explore a QAT-based approach for 1-bit quantization, also known as binarization. Motivated by the observation that the naive Round-To-Nearest (RTN) quantization method used in prior QAT works, such as LLM-QAT \cite{liu2023llm}, leads to substantial performance degradation at the 1-bit level, this approach introduced scaling factors for both input and output dimensions to mitigate the precision loss caused by binarization. These scaling factors, along with the binarized weight matrix, are initialized using a singular value decomposition and are optimized through knowledge distillation in the same manner as LLM-QAT \cite{liu2023llm}.
	
	While these QAT methods effectively mitigate the accuracy degradation caused by quantization, they are computationally expensive, as they require updating all the weights in LLMs. This challenge encouraged researchers to develop new methods aimed at preserving the beneficial adjustments from retraining while reducing the time, memory, and data requirements of the retraining process. Some of the methods which aim to achieve such a tradeoff are reviewed in Appendix \ref{s:efficient-qat}.
	
	\subsection{Improving the efficiency of QAT methods}
	\label{s:efficient-qat}
	A natural approach to bridging the gap between QAT and PTQ is to update only a small set of auxiliary parameters instead of retraining all the original weights. This strategy aims to retain the high accuracy of QAT while benefiting from the speed, memory efficiency, and reduced overfitting risks of PTQ when working with limited data. A promising choice for these auxiliary parameters is the set of quantization parameters such as scaling factors, clipping thresholds, and, zero points (if applicable). These quantization parameters (explained in Appendix \ref{sec:rtn_example}) are typically predetermined using heuristics in most PTQ methods.
	
	OmniQuant \cite{shao2023omniquant} freezes the original full-precision weights and only trains a few learnable quantization parameters which determine the clipping thresholds. Additionally, it quantizes the activations and the key-value cache by jointly learning a transformation that shifts the quantization difficulty from activations to weights. As a result, this method was able to shorten the training time from 90 hours in LLM-QAT \cite{liu2023llm} to only 1.6 hours for LLaMA-7B on a single NVIDIA A100 GPU. Also, it only used 128 segments for retraining but still outperformed existing methods by a large margin.
	
	EfficientQAT \cite{chen2024efficientqat} aims to combine the efficiency of partial training with the accuracy benefits of full retraining. To achieve this, it introduces a two-stage framework: (1) block-wise training of all parameters and (2) end-to-end fine-tuning of quantization parameters.
	In the first stage, both model and quantization parameters are trained block-by-block using reconstruction loss, enabling accurate calibration with reduced memory usage. In the second stage, the quantized weights are fixed, and only the step sizes are fine-tuned on target datasets to improve inter-block interaction. This two-stage design makes EfficientQAT a memory-efficient QAT method, as demonstrated by benchmark comparisons \cite{chen2024efficientqat}.

	\section{Comparison and replication of existing methods}
	\subsection{Comparison of existing methods}
	In this section, we compare the performance of quantized models using the methods reviewed in Section \ref{sec:intro}. Our focus is solely on how well each method preserves the accuracy of the original full-precision model. Comparisons involving memory usage and inference speed are beyond the scope of this study.
	
	As noted in Section \ref{sec:intro_ultra_low_bit}, in this study, we specifically focus on their performance under 4-bit and lower precisions. All results in Tables \ref{tab:llama_comparison} to Table \ref{tab:llama2_accuracy_comparison} are gathered from the original papers and presented in a comparable manner.
	
	In Tables \ref{tab:llama_comparison} -- \ref{tab:llama2_accuracy_comparison}, g128 indicates that the group size is 128, while W4A4 signifies that both weights and activation values are quantized to 4-bit precision. The results are sorted according to the performance on the 7B model evaluated on Wiki-Text2 dataset. The results for the models listed in Table \ref{tab:llama_comparison}--\ref{tab:llama2_accuracy_comparison} are taken directly from the original paper (where each model is discussed), and therefore the models used and the evaluation metrics vary across studies. As a result, Tables \ref{tab:llama_comparison} to \ref{tab:llama2_accuracy_comparison} include different sets of methods, depending on what performance measures were reported in each paper.
	
	\begin{table}[ht]
		\centering
		\caption{Perplexity comparison of different quantization methods evaluated on LLaMA-1-7B and LLaMA-1-13B}
		\label{tab:llama_comparison}
		\resizebox{\textwidth}{!}{
			\begin{tabular}{llr|rr|rr}
				\toprule
				Method & Bit & Additional Info & \multicolumn{2}{c|}{LLaMA-1-7B} & \multicolumn{2}{c}{LLaMA-1-13B} \\
				& & & WikiText2 & C4 & WikiText2 & C4 \\
				\midrule
				Original &  16 &               &  5.68 &  7.08 &  5.09 &  6.61 \\
				SqueezeLLM &  3 &   0.45\% outlier retention &  6.13 &  7.56 &  5.45 &  6.92 \\
				SqueezeLLM &  3 &     no outlier retention &  6.32 &  7.75 &  5.60 &  7.08 \\
				AWQ &  3 &             g128 &  6.46 &  7.92 &  5.51 &  7.07 \\
				GPTQ &  3 &             g128 &  6.55 &  7.85 &  5.62 &  7.10 \\
				RTN &  3 &             g128 &  7.01 &  8.62 &  5.88 &  7.49 \\
				DB-LLM &  2 &             g64 &  7.59 &  9.74 &  6.35 &  8.42 \\
				BinaryMoS &  1 &               &  7.97 &  9.72 &  7.16 &  8.81 \\
				GPTQ &  3 &              c &  8.06 &  9.49 &  6.76 &  8.16 \\
				OmniQuant &  2 &             g64 &  8.90 &  11.78 &  7.34 &  9.75 \\
				OmniQuant &  2 &             g128 &  9.72 &  12.97 &  7.93 &  10.36 \\
				OneBit &  1 &               &  10.38 &  11.56 &  9.18 &  10.25 \\
				OmniQuant & W4A4 &               &  11.26 &  14.51 &  10.87 &  13.78 \\
				AWQ &  3 &              c &  11.88 &  13.26 &  7.45 &  9.13 \\
				OmniQuant &  2 &              c &  15.47 &  24.89 &  13.21 &  18.31 \\
				GPTQ &  2 &             g64 &  22.10 &  17.71 &  10.06 &  11.70 \\
				SmoothQuant & W4A4 &               &  25.25 &  32.32 &  40.05 &  47.18 \\
				RTN &  3 &              c &  25.73 &  28.26 &  11.39 &  13.22 \\
				BiLLM &  1 &               &  41.66 &  48.15 &  14.56 &  16.67 \\
				GPTQ &  2 &             g128 &  44.01 &  27.71 &  15.60 &  15.29 \\
				RTN &  2 &             g64 & 188.32 & 151.43 & 101.87 &  76.00 \\
				PB-LLM &  1 & 10\% salient weights retained & 198.37 & 157.35 &  35.83 &  39.79 \\
				RTN &  2 &             g128 & 1.9e+03 & 1.0e+03 & 781.20 & 447.64 \\
				GPTQ &  2 &              c & 2.1e+03 & 689.13 & 5.5e+03 & 2.5e+03 \\
				RTN &  2 &              c & 1.1e+05 & 1.3e+05 & 6.8e+04 & 5.6e+04 \\
				AWQ &  2 &             g64 & 2.5e+05 & 2.8e+05 & 2.7e+05 & 2.2e+05 \\
				AWQ &  2 &             g128 & 2.6e+05 & 1.9e+05 & 2.8e+05 & 2.3e+05 \\
				\bottomrule
			\end{tabular}
		}
	\end{table}
	
	In Table \ref{tab:llama_comparison}, we present a comparison of the performance of several methods discussed in Section \ref{sec:intro}, evaluated on two models from LLaMA-1 \cite{touvron2023llama}. The evaluation is based on perplexity measured on two commonly used datasets: WikiText-2 \cite{merity2017pointer} and C4 \cite{raffel2020exploring}. Perplexity is a commonly used metric to evaluate the performance of a language model. It is defined as the exponentiated average negative log-likelihood of a given token sequence. Formally, for a sequence of tokens \( X = (x_1, x_2, \dots, x_N) \), the perplexity is defined as:
	
	\begin{equation}
		\text{PPL}(X) = \exp \left( -\frac{1}{N} \sum_{i=1}^{N} \log P(x_i \mid x_1, x_2, \dots, x_{i-1}) \right)
	\end{equation}
	
	Here, \( P(x_i \mid x_1, x_2, \dots, x_{i-1}) \) denotes the probability assigned by the language model to the token \( x_i \), given its preceding context. Lower perplexity indicates that the model predicts the sequence more accurately.
	
	\begin{table}[ht]
		\centering
		\caption{Zero-shot accuracy comparison of different quantization methods evaluated on LLaMA-1-7B and LLaMA-1-13B}
		\label{tab:llama1_accuracy_comparison}
		
		\begin{subtable}{\textwidth}
			\centering
			\caption{LLaMA-1-7B}
			\resizebox{\textwidth}{!}{ 
				\begin{tabular}{llr|rrrrrr}
					\toprule
					Method & Bit & Additional info & WinoG & Hellaswag & PIQA & BoolQ & ARC-e & ARC-c \\
					\midrule
					Original & 16 &               & 66.85 & 72.99 & 77.37 & 73.21 & 52.53 & 41.38 \\
					DB-LLM & 2 &             g64 & 61.01 & 60.71 & 72.14 &    & 44.70 & 33.62 \\
					OneBit & 1 &               & 60.30 & 50.73 & 67.46 & 62.51 & 41.71 & 29.61 \\
					BinaryMoS & 1 &               & 58.88 & 58.18 & 71.82 & 64.59 & 42.09 & 31.31 \\
					BiLLM & 1 &               & 51.14 & 34.64 & 58.65 & 62.23 & 33.08 & 25.68 \\
					PB-LLM & 1 & 10\% salient weights retained & 49.17 & 27.23 & 53.53 & 60.51 & 27.48 & 26.02 \\
					\bottomrule
				\end{tabular}
			}
		\end{subtable}
		
		\vspace{1em}
		
		\begin{subtable}{\textwidth}
			\centering
			\caption{LLaMA-1-13B}
			\resizebox{\textwidth}{!}{ 
				\begin{tabular}{llr|rrrrrr}
					\toprule
					Method & Bit & Additional info & WinoG & Hellaswag & PIQA & BoolQ & ARC-e & ARC-c \\
					\midrule
					Original & 16 &               & 70.17 & 76.24 & 79.05 & 68.47 & 59.85 & 44.54 \\
					DB-LLM & 2 &             g64 & 64.72 & 68.29 & 74.16 &    & 51.18 & 37.54 \\
					OneBit & 1 &               & 62.90 & 56.78 & 70.67 & 64.16 & 44.53 & 32.10 \\
					BinaryMoS & 1 &               & 60.93 & 64.05 & 73.88 & 63.82 & 44.28 & 33.11 \\
					BiLLM & 1 &               & 59.43 & 52.24 & 68.17 & 62.53 & 41.91 & 29.94 \\
					PB-LLM & 1 & 10\% salient weights retained & 52.17 & 33.97 & 58.70 & 62.17 & 31.86 & 23.63 \\
					\bottomrule
				\end{tabular}
			}
		\end{subtable}
		
	\end{table}

	In Table \ref{tab:llama1_accuracy_comparison}, we use the same two Llama-1 models to present a comparison of some of the ultra-low-bit quantization methods based on various common sense reasoning tasks such as BoolQ \cite{clark2019boolq}, WinoGrande \cite{sakaguchi2021winogrande}, PIQA \cite{bisk2020piqa}, HellaSwag \cite{zellers2019hellaswag}, Arc-Easy \cite{clark2018think}, and Arc-Challenge \cite{clark2018think}. These datasets consist of multiple-choice questions from various domains. For each dataset, zero-shot accuracy, which is the accuracy without task-specific fine-tuning is reported.

	The perplexity comparison for two models from the LLaMA-2 family \cite{touvron2023llama2} are provided in Table \ref{tab:llama2_comparison}. Table \ref{tab:llama2_comparison} shows that the accuracy drop for 3-bit quantized models has become less pronounced. However, for 2-bit and 1-bit quantization, even QAT-based methods continue to exhibit a relatively large performance gap compared to the original model. Note that large performance degradation is observed for models such as DB-LLM \cite{chen2024db}, BinaryMoS \cite{jo2024mixture}, and OneBit \cite{xu2024onebit}, that are specifically designed for low-bit quantization. 
	
	The accuracy comparison for LLaMA-2-7B and LLaMA-2-13B models is provided in Table \ref{tab:llama2_accuracy_comparison}. It indicates that there remains an accuracy gap of approximately 5\% to 10\% to be potentially bridged by better models.

	\begin{table}
		\centering
		\caption{Perplexity comparison of different quantization methods evaluated on LLaMA-2-7B and LLaMA-2-13B}
		\label{tab:llama2_comparison}
		\resizebox{\textwidth}{!}{
			\begin{tabular}{llr|rr|rr}
				\toprule
				Method & Bit & Additional Info & \multicolumn{2}{c|}{LLaMA-2-7B} & \multicolumn{2}{c}{LLaMA-2-13B} \\
				& & & WikiText2 & C4 & WikiText2 & C4 \\
				\midrule
				Original &  16 &               &  5.47 &  6.97 &  4.88 &  6.46 \\
				ApiQ &  3 &          block-wise &  5.77 &  7.48 &  5.12 &  6.83 \\
				EfficientQAT &  3 &             g128 &  5.81 &  7.34 &  5.12 &  6.73 \\
				SqueezeLLM &  3 &   0.45\% outlier retention &  5.96 &  7.51 &  5.23 &  6.82 \\
				SqueezeLLM &  3 &     no outlier retention &  6.18 &  7.72 &  5.36 &  6.97 \\
				AWQ &  3 &             g128 &  6.24 &  7.84 &  5.32 &  6.94 \\
				GPTQ &  3 &             g128 &  6.29 &  7.89 &  5.42 &  7.00 \\
				RTN &  3 &             g128 &  6.66 &  8.40 &  5.51 &  7.18 \\
				EfficientQAT &  2 &             g64 &  6.86 &  8.50 &  5.96 &  7.59 \\
				EfficientQAT &  2 &             g128 &  7.19 &  8.79 &  6.08 &  7.75 \\
				DB-LLM &  2 &             g64 &  7.23 &  9.62 &  6.19 &  8.38 \\
				ApiQ &  2 &          block-wise &  7.59 &  10.56 &  6.44 &  8.93 \\
				BinaryMoS &  1 &               &  7.88 &  9.75 &  7.08 &  8.91 \\
				GPTQ &  3 &              c &  8.37 &  9.81 &  6.44 &  8.02 \\
				OmniQuant &  2 &             g64 &  9.62 &  12.72 &  7.56 &  10.05 \\
				OneBit &  1 &               &  9.73 &  11.11 &  8.76 &  10.15 \\
				LoftQ &  3 &               &  10.72 &  12.79 &  6.89 &  8.72 \\
				OmniQuant &  2 &             g128 &  11.06 &  15.02 &  8.26 &  11.05 \\
				OmniQuant & W4A4 &               &  14.26 &  18.02 &  12.30 &  14.55 \\
				GPTQ &  2 &             g64 &  20.85 &  19.40 &  22.44 &  12.48 \\
				AWQ &  3 &              c &  24.00 &  23.85 &  10.45 &  13.07 \\
				BiLLM &  1 &               &  27.72 &  36.34 &  20.71 &  27.19 \\
				GPTQ &  2 &             g128 &  36.77 &  33.70 &  28.14 &  20.97 \\
				OmniQuant &  2 &              c &  37.37 &  90.64 &  17.21 &  26.76 \\
				PB-LLM &  1 & 10\% salient weights retained &  76.75 &  85.92 & 155.25 & 151.15 \\
				SmoothQuant & W4A4 &               &  83.12 &  77.27 &  35.88 &  43.19 \\
				RTN &  2 &             g64 & 431.97 & 475.35 &  26.22 &  28.69 \\
				RTN &  3 &              c & 539.48 & 402.35 &  10.68 &  12.51 \\
				LoftQ &  2 &               & 1.0e+03 & 670.00 &  59.94 &  72.64 \\
				RTN &  2 &             g128 & 4.2e+03 & 4.9e+03 & 122.08 & 139.65 \\
				GPTQ &  2 &              c & 7.7e+03 &     & 2.1e+03 & 323.12 \\
				RTN &  2 &              c & 3.8e+04 & 4.8e+04 & 5.6e+04 & 7.2e+04 \\
				AWQ &  2 &             g64 & 2.1e+05 & 1.6e+05 & 1.2e+05 & 9.5e+04 \\
				AWQ &  2 &             g128 & 2.2e+05 & 1.7e+05 & 1.2e+05 & 9.4e+04 \\
				\bottomrule
			\end{tabular}
		}
	\end{table}
	
	\begin{table}
		\centering
		\caption{Zero-shot accuracy comparison of different quantization methods evaluated on LLaMA-2-7B and LLaMA-2-13B}
		\label{tab:llama2_accuracy_comparison}
		
		\begin{subtable}{\textwidth}
			\centering
			\caption{LLaMA-2-7B}
			\resizebox{\textwidth}{!}{
				\begin{tabular}{llr|rrrrrr}
					\toprule
					Method & Bit & Additional info & WinoG & Hellaswag & PIQA & BoolQ & ARC-e & ARC-c \\
					\midrule
					Original & 16 &           & 67.09 & 72.94 & 76.88 & 71.10 & 53.58 & 40.61 \\
					DB-LLM  & 2 & g64         & 61.72 & 61.98 & 73.18 &    & 45.20 & 33.53 \\
					OneBit  & 1 &           & 58.41 & 52.58 & 68.12 & 63.06 & 41.58 & 29.61 \\
					BinaryMoS& 1 &           & 56.18 & 59.41 & 71.55 & 65.02 & 41.84 & 30.03 \\
					BiLLM  & 1 &           & 53.11 & 35.18 & 59.19 & 62.14 & 34.22 & 26.54 \\
					PB-LLM  & 1 & 10\% salient weights retained
					& 50.11 & 26.87 & 52.82 & 62.17 & 26.89 & 24.31 \\
					\bottomrule
				\end{tabular}
			}
		\end{subtable}
		
		\vspace{1em}
		
		\begin{subtable}{\textwidth}
			\centering
			\caption{LLaMA-2-13B}
			\resizebox{\textwidth}{!}{
				\begin{tabular}{llr|rrrrrr}
					\toprule
					Method & Bit & Additional info & WinoG & Hellaswag & PIQA & BoolQ & ARC-e & ARC-c \\
					\midrule
					Original & 16 &           & 69.77 & 76.62 & 79.05 & 68.99 & 57.95 & 44.20 \\
					DB-LLM  & 2 & g64         & 64.09 & 68.04 & 75.14 &    & 51.64 & 38.14 \\
					OneBit  & 1 &           & 61.72 & 56.43 & 70.13 & 65.20 & 43.10 & 33.62 \\
					BinaryMoS& 1 &           & 58.98 & 63.80 & 73.12 & 66.12 & 45.71 & 33.19 \\
					BiLLM  & 1 &           & 56.35 & 38.05 & 62.51 & 62.20 & 4.69 & 27.73 \\
					PB-LLM  & 1 & 10\% salient weights retained
					& 49.48 & 28.89 & 53.26 & 37.82 & 28.28 & 23.72 \\
					\bottomrule
				\end{tabular}
			}
		\end{subtable}
		
	\end{table}
	
	\FloatBarrier

	\subsection{Replication study of ApiQ}
\begin{figure}[ht]
		\centering
		\begin{subfigure}[b]{0.49\linewidth}
			\centering
			\includegraphics[width=\linewidth]{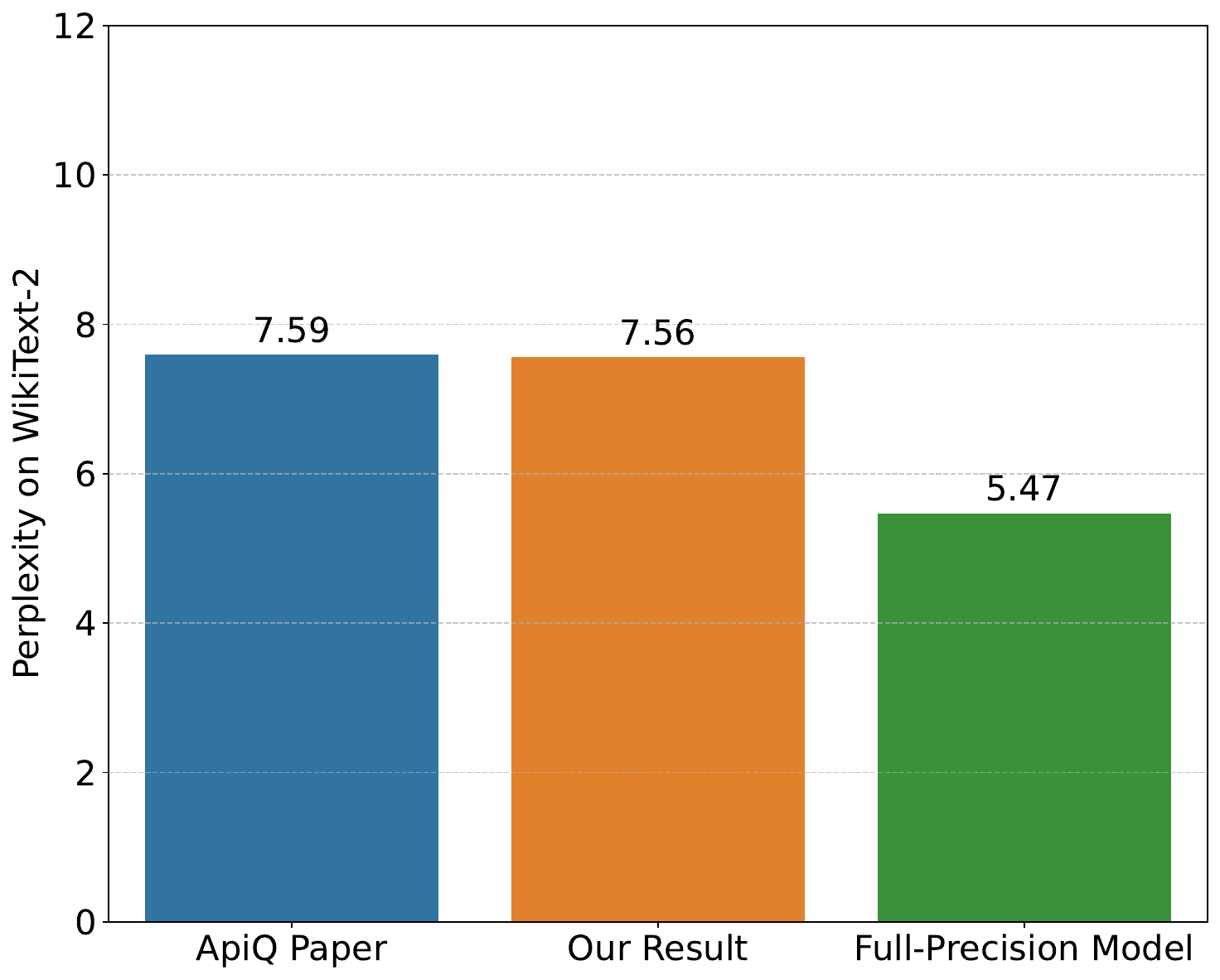}
			\caption{WikiText-2}
			\label{fig:reproduced_ppl_wikitext2}
		\end{subfigure}
		\hfill
		\begin{subfigure}[b]{0.49\linewidth}
			\centering
			\includegraphics[width=\linewidth]{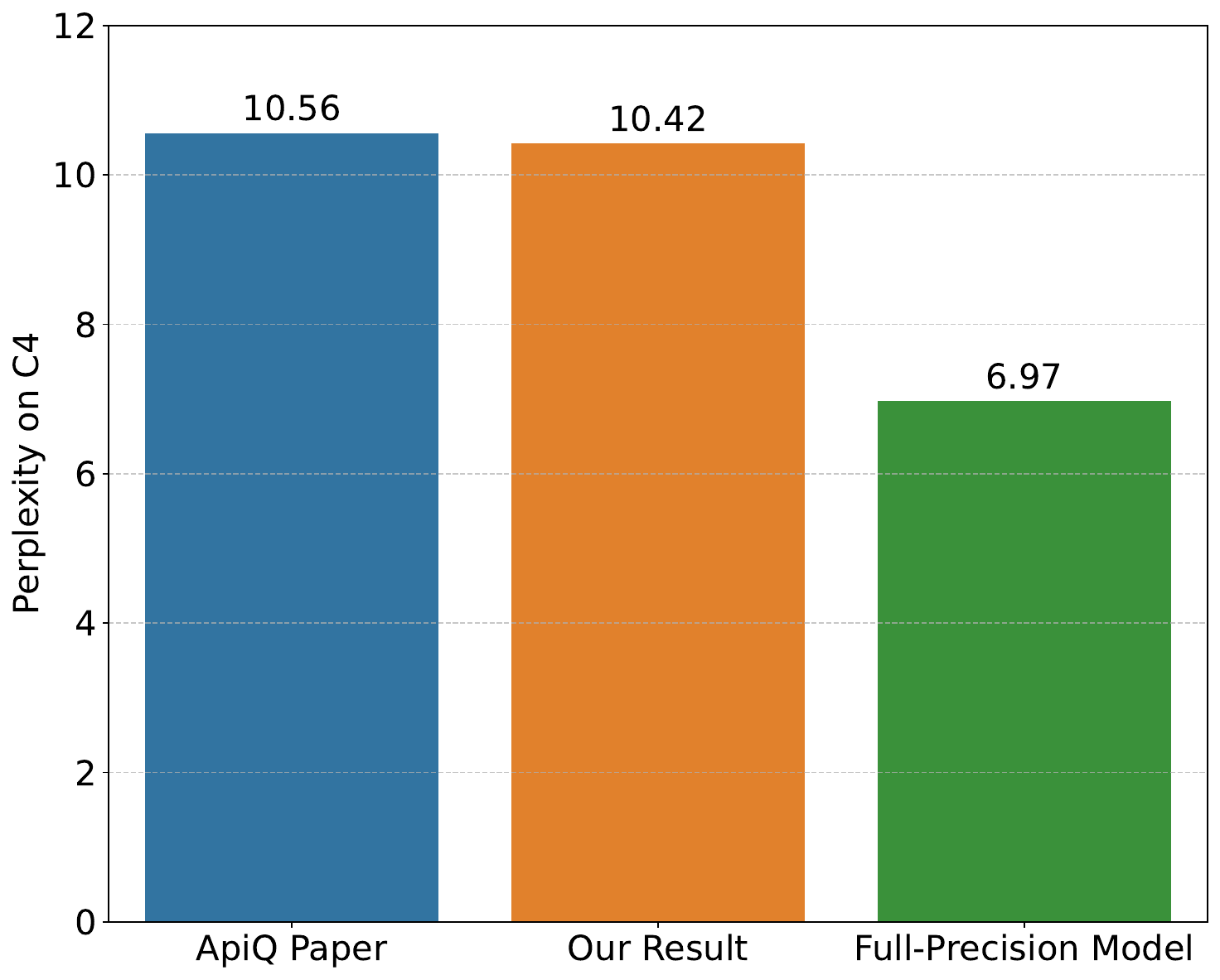}
			\caption{C4}
			\label{fig:reproduced_ppl_c4}
		\end{subfigure}
		\caption{Perplexity comparison between our reproduced results and the original results reported for ApiQ (for 2-bit quantization of LLaMA-2-7B model). Performance of the full-precision model is also added for reference.}
		\label{fig:reproduced_results}
	\end{figure}

	We replicated the results from ApiQ \cite{liao2024apiq} so that we have a baseline for measuring the effects of our further modifications. We used the same hyperparameter sets as described in their paper and outlined in Table \ref{tab:hyperparameters} and ran quantization on LLaMA-2-7B model. The main performance results are presented in Fig.\ \ref{fig:reproduced_results}. Although minor differences were observed, the results we obtained were largely consistent with those reported in the original paper \cite{liao2024apiq}.

	\section{Evaluation of preliminary ideas}
	\label{sec:result_preliminary_ideas}
	\subsection{Performance of ApiQ + BinaryMoS}\label{sec:ApiQ+BinaryMoS_results}
\begin{figure}[ht]
		\centering
		\includegraphics[width=0.55\linewidth]{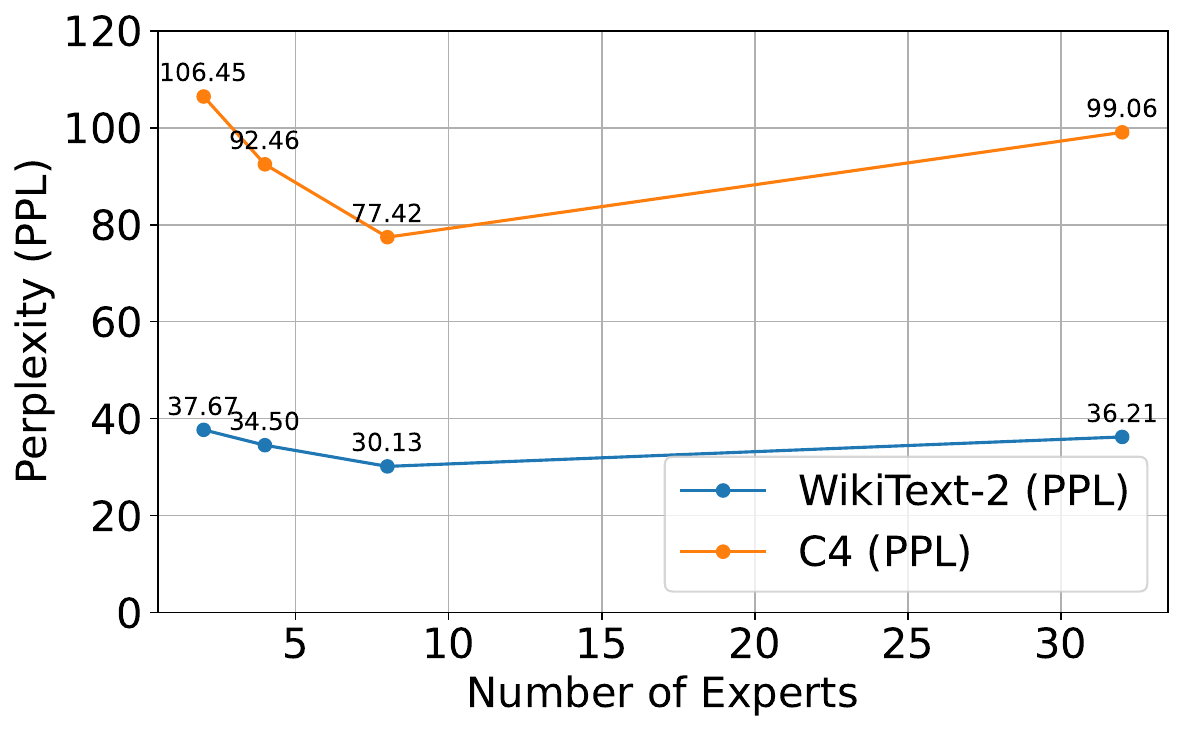}
		\caption{Performance of ApiQ+BinaryMoS on LLaMA-2-7B model with different number of experts}
		\label{fig:ApiQ+BinaryMoS_with_different_experts}
	\end{figure}

	In this experiment, we evaluated the approach described in Section \ref{sec:integration_of_binarymos}. The hyperparameters used were identical to those listed in Table \ref{tab:hyperparameters}, except for the learning rate of the quantization parameters, which was adjusted to 5e-4 as it yielded better results. We tested different numbers of scaling factors (i.e. experts) and the results are shown in Fig.\ \ref{fig:ApiQ+BinaryMoS_with_different_experts}. Despite extensive hyperparameter tuning, our approach did not yield results superior to those presented in the third row of Fig.\ \ref{fig:ApiQ+BinaryMoS_with_different_experts}, which are substantially worse than the results reported in the original BinaryMoS paper. The key distinction between our configuration and their original setup lies in the treatment of trainable parameters. While the original paper trained both weights and quantization parameters jointly, our approach froze the weights, introduced LoRA adapters, and trained only the quantization parameters and LoRA weights. This observation highlights a fundamental limitation in the capacity of LoRA weights to mitigate the accuracy degradation caused by quantization, particularly under the stringent condition of 1-bit quantization.
	
	\FloatBarrier
	
	\subsection{Performance of ApiQ + DB-LLM}\label{sec:preliminary_exp_db_llm}
	In this experiment, we evaluated the approach described in Section \ref{sec:integration_of_db_llm}.
	We initially evaluated our proposed method using the same set of hyperparameters as ApiQ (as specified in Table \ref{tab:hyperparameters}), except for the learning rate of the quantization parameters. For this, we adopted a value of 1e-4, as our hyperparameter tuning experiments indicated that it leads to more stable training. Since this naive integration led to substantial overfitting and degraded performance, we also tested the saliency-aware weight preservation regularization proposed in Section \ref{sec:proposed_methods}. We exclusively evaluated the after-LoRA variant, as it demonstrated superior performance when tested using the original ApiQ method. The results are presented in Tables \ref{tab:ApiQ_DB-LLM_experimental_results} -- \ref{tab:ApiQ+DB_LLM_saliency_aware_weight_preservation_reg}. While the inclusion of the regularization term substantially improved performance, our method did not surpass the original ApiQ approach when we used the same amount of training data as the original ApiQ. However, as the third line of Table \ref{tab:ApiQ+DB_LLM_saliency_aware_weight_preservation_reg} indicates, using more training data can improve the performance decently, surpassing that of the original ApiQ method.
	
	We observed that directly using the quantization formulation of DB-LLM, which was originally proposed as a QAT method, can reduce quantization loss. However, without large and diverse training data, addressing the issue of overfitting remains challenging.
	
	\begin{table}
		\centering
		\caption{Performance of ApiQ + DB-LLM on LLaMA-2-7B model}
		\label{tab:ApiQ_DB-LLM_experimental_results}
		\resizebox{\textwidth}{!}{%
			\begin{tabular}{c c c c c c c c c}
				\toprule
				\multirow{2}{*}{\textbf{Method}} 
				& \multicolumn{2}{c}{\textbf{PPL}\,$\downarrow$}
				& \multicolumn{6}{c}{\textbf{Accuracy (\%)}\,$\uparrow$} \\
				\cmidrule(lr){2-3}
				\cmidrule(lr){4-9}
				& \textbf{WikiText2} & \textbf{C4}
				& \textbf{WinoGrande} & \textbf{HellaSwag} 
				& \textbf{ArcC}    & \textbf{ArcE}
				& \textbf{PiQA}    & \textbf{Average} \\
				\midrule
				\textbf{Full-Precision Model} & 5.47 & 6.97 & 69.22 & 57.16 & 43.52 & 76.26 & 78.07 & 64.85 \\
				\textbf{Original ApiQ} & 7.56 & 10.42 & 62.83 & 46.53 & 31.57 & 66.54 & 72.09 & 55.91 \\
				\textbf{ApiQ + DB-LLM} & 8.26 & 12.95 & 60.22 & 44.64 & 20.90 & 37.37 & 59.98 & 44.54 \\
				\bottomrule
			\end{tabular}
		}
	\end{table}
	
	\begin{table}[ht]
		\centering
		\caption{Performance of ApiQ + DB-LLM with saliency-aware weight preservation regularization on LLaMA-2-7B (After-LoRA only). Dataset: training set; Coefficient: regularization weight; Samples: number of training examples; Epochs: training epochs per layer.}
		\label{tab:ApiQ+DB_LLM_saliency_aware_weight_preservation_reg}
		\resizebox{\linewidth}{!}{%
			\begin{tabular}{l c c c c c c c c c c c}
				\toprule
				\multirow{2}{*}{\textbf{Dataset}}
				& \multirow{2}{*}{\textbf{Coefficient}}
				& \multirow{2}{*}{\textbf{Samples}}
				& \multirow{2}{*}{\textbf{Epochs}}
				& \multicolumn{2}{c}{\textbf{PPL}\,$\downarrow$}
				& \multicolumn{6}{c}{\textbf{Accuracy (\%)}\,$\uparrow$} \\
				\cmidrule(lr){5-6}
				\cmidrule(lr){7-12}
				& & & 
				& \textbf{WikiText2} & \textbf{C4}
				& \textbf{WinoGrande} & \textbf{HellaSwag} 
				& \textbf{ArcC}    & \textbf{ArcE}
				& \textbf{PiQA}    & \textbf{Average} \\
				\midrule
				WikiText-2 & 10 & 128 & 20 & 7.52 & 10.38 & 62.35 & 46.26 & 31.40 & 66.37 & 71.55 & 55.58 \\
				C4     & 10 & 128 & 20 & 8.39 & 10.75 & 62.83 & 46.85 & 31.31 & 66.25 & 71.44 & 55.73 \\
				C4     & 10 & 512 & 10 & 7.46 & 10.43 & 62.35 & 46.72 & 33.45 & 68.14 & 72.09 & 56.55 \\
				\bottomrule
			\end{tabular}%
		}
	\end{table}

	\FloatBarrier

	\subsection{Lessons learned from combining QAT methods with ApiQ}
	\label{sec:discussion_QAT_plus_ApiQ}
	
	We investigated the idea of replacing ApiQ’s quantized linear layers with those from BinaryMoS and DB-LLM. Each approach originally achieves strong performance when full model weights are retrained on ample data, but we hypothesized that LoRA-based partial finetuning might be sufficient to close the gap between a compressed model and its full-precision counterpart. However, the results indicated otherwise: freezing the original weights while training only LoRA adapters and quantization parameters did not yield accuracy improvements beyond baseline ApiQ. For instance, BinaryMoS’s multi-expert binarization, effective under standard QAT, still struggled when only LoRA weights were permitted to be updated. On the other hand, DB-LLM’s flexible dual binarization helped reduce quantization loss but showed substantial overfitting to the limited calibration data, preventing it from outperforming ApiQ’s simpler design.
	
	Our result of integrating BinaryMoS into ApiQ suggests that the effectiveness of QAT-based strategies in the 1-bit regime is highly dependent on the ability to retrain a substantial portion of the model’s weights. While ApiQ demonstrated that LoRA adapters can effectively compensate for quantization errors in the 2-bit setting, this compensatory effect does not extend to 1-bit scenarios, where the representational capacity of quantized weights is more severely constrained. Notably, our results with BinaryMoS underperformed not only BinaryMoS’s original implementation but also its predecessor, OneBit \cite{xu2024onebit}, which uses a simpler and thus less expressive quantization scheme with only a single pair of scaling factors. This comparison underscores a key insight: increasing the expressiveness of quantization alone is insufficient when the main weights remain frozen. Instead, retraining the original weights remains essential for recovering performance under 1-bit quantization.

	In contrast, our results of integrating DB-LLM into ApiQ suggest a different implication. Its quantization formulation is sufficiently expressive—evidenced by its tendency to overfit the limited calibration data—indicating that the availability of a large and diverse training set is equally critical. This observation aligns with findings from EfficientQAT \cite{chen2024efficientqat}, whose ablation studies showed that when both weights and quantization parameters (scaling factors and zero points) are trained in a block-wise manner, at least 4096 samples are required to mitigate overfitting. Although our method freezes the original weights and trains fewer parameters by introducing LoRA adapters, our results similarly indicate that 128 samples (the number we used) are not enough to prevent overfitting. The third row of Table \ref{tab:ApiQ+DB_LLM_saliency_aware_weight_preservation_reg} demonstrates that increasing the volume of calibration data consistently improves performance beyond the ApiQ baseline. This highlights both the risk of overfitting and the imperative of employing larger datasets when using more expressive quantization schemes.

	\section{Additional results on the naive regularizer}
	\label{sec:result_reg_original_weights_preservation}
	

	We evaluated the effectiveness of the naive weight preservation regularizer introduced in Section \ref{ss:reg2}.

	\begin{table}
		\centering
		\caption{Performance of ApiQ with weight preservation regularization on LLaMA-2-7B model}
		\label{tab:ApiQ_weight_preservation_reg}
		\resizebox{\textwidth}{!}{%
			\begin{tabular}{l l c c c c c c c c}
				\toprule
				\multirow{2}{*}{\textbf{LoRA Variant|}} 
				\multirow{2}{*}{\textbf{Coefficient}}
				& \multicolumn{2}{r}{\textbf{PPL}\,$\downarrow$}
				& \multicolumn{6}{c}{\textbf{Accuracy (\%)}\,$\uparrow$} \\
				\cmidrule(lr){3-4} 
				\cmidrule(lr){5-10}
				& & \textbf{WikiText2} & \textbf{C4}
				& \textbf{WinoGrande} & \textbf{HellaSwag} 
				& \textbf{ArcC}    & \textbf{ArcE}
				& \textbf{PiQA}    & \textbf{Average} \\
				\midrule
				\multicolumn{2}{c}{\textbf{Full-Precision Model}} & 5.47 & 6.97 & 69.22 & 57.16 & 43.52 & 76.26 & 78.07 & 64.85 \\
				\multicolumn{2}{c}{\textbf{Original ApiQ (No regularization)}} & 7.56 & 10.42 & 62.83 & 46.53 & 31.57 & 66.54 & 72.09 & 55.91 \\
				\midrule
				AfterLoRA & 1e-3 & 8.77 & 11.5 & 57.62 & 45.97 & 27.82 & 59.09 & 68.93 & 51.88 \\
				AfterLoRA & 1e-4 & 8.12 & 10.83 & 60.83 & 46.68 & 30.72 & 64.27 & 72.09 & 54.92 \\
				AfterLoRA & 1e-5 & 7.83 & 10.84 & 60.93 & 45.84 & 31.74 & 66.25 & 71.38 & 55.21 \\
				AfterLoRA & 1e-6 & 8.82 & 21.91 & 60.06 & 43.15 & 30.12 & 64.48 & 70.73 & 53.71 \\
				\midrule
				BeforeLoRA & 1e-3 & 8.21 & 12.16 & 61.64 & 45.03 & 30.46 & 62.29 & 70.35 & 53.95 \\
				BeforeLoRA & 1e-4 & 8.12 & 12.77 & 61.64 & 41.19 & 32.59 & 66.41 & 70.78 & 54.52 \\
				BeforeLoRA & 1e-5 & 7.85 & 16.77 & 61.40 & 43.91 & 31.91 & 66.79 & 71.06 & 55.01 \\
				BeforeLoRA & 1e-6 & 7.69 & 12.72 & 61.40 & 45.22 & 31.06 & 66.20 & 72.03 & 55.18 \\
				\bottomrule
			\end{tabular}
		}
	\end{table}
	
	We tested using different coefficients for the regularization term and also evaluated the two variants (before / after LoRA) mentioned in Section \ref{ss:reg2}. The results are shown on Table \ref{tab:ApiQ_weight_preservation_reg}. None of the results presented in this table surpassed the performance of the original ApiQ method, suggesting that the naive weight preservation approach is ineffective for regularizing the loss function.
	
	\section{Ablation studies}
	\subsection{Applying proposed regularization term under 3-bit quantization}
	To validate the robustness of the saliency-aware weight preservation term introduced in Section \ref{ss:reg3}, we extended its use to 3-bit quantization. The training settings used here follow the settings reported in ApiQ \cite{liao2024apiq}, which are provided in Table \ref{tab:hyperparameters_3bit}. We used the AfterLoRA variant and used the gradients calculated from WikiText-2 dataset to calculate the saliency values. Through hyperparameter tuning, we found that regularization stronger than the 2-bit case improved performance at 3 bits. We also observed that a coefficient of 10 yielded the best results, as indicated in Table \ref{tab:ApiQ_saliency_aware_weight_preservation_reg_3bit} in comparison with the baselines. With the proposed regularization, we were able to achieve an increase of 14 basis points in the average accuracy (the differece between second and the third rows of the rightmost column in Table \ref{tab:ApiQ_saliency_aware_weight_preservation_reg_3bit}). This increase accounts for 7.91 \% relative improvement in the accuracy degradation incurred by the ApiQ method.
	
	\begin{table}
		\centering
		\caption{Training settings for 3-bit quantization of LLaMA-2-7B model. Other hyperparameters are the same as Table \ref{tab:hyperparameters}.}
		\begin{tabular}{ll}
			\toprule
			\textbf{Hyperparameter}           & \textbf{Value} \\ 
			\midrule
			\multicolumn{2}{l}{\textbf{Optimizer and Training Settings}} \\ 
			Optimizer                  & AdamW     \\ 
			Weight decay for quantization parameters  & 0.1      \\ 
			Learning rate for quantization parameters  & 0.001     \\ 
			Weight decay for LoRA weights        & 0.0      \\ 
			Learning rate for LoRA weights       & 0.0005     \\ 
			Batch size                 & 1       \\ 
			Epochs (per layer)                  & 20       \\ 
			
			\bottomrule
		\end{tabular}
		\label{tab:hyperparameters_3bit}
	\end{table}
	
	\begin{table}
		\centering
		\caption{Performance of ApiQ with saliency-aware weight preservation regularization under 3-bit quantization evaluated on LlaMA-2-7B model}
		\label{tab:ApiQ_saliency_aware_weight_preservation_reg_3bit}
		\resizebox{\textwidth}{!}{%
			\begin{tabular}{c c c c c c c c c}
				\toprule
				\multirow{2}{*}{\textbf{Method}} 
				& \multicolumn{2}{c}{\textbf{PPL}\,$\downarrow$}
				& \multicolumn{6}{c}{\textbf{Accuracy (\%)}\,$\uparrow$} \\
				\cmidrule(lr){2-3}
				\cmidrule(lr){4-9}
				& \textbf{WikiText2} & \textbf{C4}
				& \textbf{WinoGrande} & \textbf{HellaSwag} 
				& \textbf{ArcC}    & \textbf{ArcE}
				& \textbf{PiQA}    & \textbf{Average} \\
				\midrule
				\textbf{Full-Precision Model} & 5.47 & 6.97 & 69.22 & 57.16 & 43.52 & 76.26 & 78.07 & 64.85 \\
				\textbf{Original ApiQ} & 5.98 & 7.45 & 68.22 & 55.09 & 40.44 & 73.65 & 77.37 & 63.08 \\
				\textbf{ApiQ with Proposed Regularization} & 5.78 & 7.46 & 68.19 & 55.36 & 41.21 & 73.99 & 77.37 & 63.22 \\
				\bottomrule
			\end{tabular}
		}
	\end{table}
	
	\subsection{Investigating the effects of different calibration datasets}
	To evaluate the sensitivity of our proposed method to the calibration dataset and its size, we carried out the same experiment with different calibration datasets. We conducted experiments on the Penn Treebank (PTB) dataset \cite{marcus-etal-1993-building} as well as on a composite corpus formed by equally mixing WikiText-2 \cite{merity2017pointer}, C4 \cite{raffel2020exploring}, and PTB. To ensure comparability with the results in Section \ref{sec:result_reg_saliency_aware_weights_preservation}, we sampled 128 examples from each dataset, applied a regularization coefficient of 2, and evaluated the AfterLoRA variant. All other hyperparameters remain as listed in Table \ref{tab:hyperparameters}. The results for PTB dataset and the mixed dataset are provided in Table \ref{tab:ApiQ_saliency_aware_weight_preservation_reg_ptb} and Table \ref{tab:ApiQ_saliency_aware_weight_preservation_reg_mix} respectively.
	
	When calibrated on PTB, our method achieved a gain of 115 basis points in average accuracy (corresponding to the difference between the second and third rows of the rightmost column in Table \ref{tab:ApiQ_saliency_aware_weight_preservation_reg_ptb}). This gain accounts for a relative improvement of 11.83 \% in the accuracy degradation incurred by ApiQ. Using the mixed dataset, the proposed regularization delivered an accuracy improvement of 203 basis points (the difference between the second and third rows of the rightmost column in Table \ref{tab:ApiQ_saliency_aware_weight_preservation_reg_mix}). This accounts for 17.29 \% relative improvement in accuracy degradation incurred by ApiQ. These results indicate that, although the baseline ApiQ’s performance depends on the calibration data, our approach consistently enhances accuracy irrespective of the dataset chosen. Furthermore, we observe that the improvement is particularly pronounced on datasets where the accuracy of the original ApiQ method is low, yielding greater relative gains in such challenging cases.
	
	\begin{table}
		\centering
		\caption{Performance of ApiQ with Saliency-Aware Weight Preservation Regularization on the LLaMA-2-7B Model Calibrated Using the PTB Dataset}
		\label{tab:ApiQ_saliency_aware_weight_preservation_reg_ptb}
		\resizebox{\textwidth}{!}{%
			\begin{tabular}{c c c c c c c c c}
				\toprule
				\multirow{2}{*}{\textbf{Method}} 
				& \multicolumn{2}{c}{\textbf{PPL}\,$\downarrow$}
				& \multicolumn{6}{c}{\textbf{Accuracy (\%)}\,$\uparrow$} \\
				\cmidrule(lr){2-3}
				\cmidrule(lr){4-9}
				& \textbf{WikiText2} & \textbf{C4}
				& \textbf{WinoGrande} & \textbf{HellaSwag} 
				& \textbf{ArcC}    & \textbf{ArcE}
				& \textbf{PiQA}    & \textbf{Average} \\
				\midrule
				\textbf{Full-Precision Model} & 5.47 & 6.97 & 69.22 & 57.16 & 43.52 & 76.26 & 78.07 & 64.85 \\
				\textbf{Original ApiQ} & 9.22 & NaN & 62.19 & 46.46 & 31.06 & 64.31 & 71.60 & 55.13 \\
				\textbf{ApiQ with Proposed Regularization} & 8.40 & 10.47 & 63.61 & 47.20 & 31.57 & 66.67 & 72.36 & 56.28 \\
				\bottomrule
			\end{tabular}
		}
	\end{table}

	\begin{table}
		\centering
		\caption{Performance of ApiQ with saliency-aware weight preservation regularization on the LLaMA-2-7B model calibrated using the mixed dataset}
		\label{tab:ApiQ_saliency_aware_weight_preservation_reg_mix}
		\resizebox{\textwidth}{!}{%
			\begin{tabular}{c c c c c c c c c}
				\toprule
				\multirow{2}{*}{\textbf{Method}} 
				& \multicolumn{2}{c}{\textbf{PPL}\,$\downarrow$}
				& \multicolumn{6}{c}{\textbf{Accuracy (\%)}\,$\uparrow$} \\
				\cmidrule(lr){2-3}
				\cmidrule(lr){4-9}
				& \textbf{WikiText2} & \textbf{C4}
				& \textbf{WinoGrande} & \textbf{HellaSwag} 
				& \textbf{ArcC}    & \textbf{ArcE}
				& \textbf{PiQA}    & \textbf{Average} \\
				\midrule
				\textbf{Full-Precision Model} & 5.47 & 6.97 & 69.22 & 57.16 & 43.52 & 76.26 & 78.07 & 64.85 \\
				\textbf{Original ApiQ} & NaN & NaN & 60.62 & 44.40 & 29.86 & 59.89 & 70.78 & 53.11 \\
				\textbf{ApiQ with Proposed Regularization} & 8.41 & 11.01 & 63.93 & 45.61 & 30.12 & 64.39 & 71.65 & 55.14 \\
				\bottomrule
			\end{tabular}
		}
	\end{table}

		\subsection{Investigating the effect of calibration dataset size}
	
	To investigate the effect of calibration dataset size on the accuracy of the quantized model, we used WikiText-2 as the calibration dataset. We increased the number of training samples from 128 to 512 and trained each layer for five epochs, which proved more effective with the larger dataset. We set the regularization coefficient to 2, which yielded the best performance as shown in Table \ref{tab:ApiQ_saliency_aware_weight_preservation_reg_coeffs}. All other hyperparameters remained the same as those listed in Table \ref{tab:hyperparameters}.
	
		\begin{table}[ht!]
		\centering
		\caption{Performance comparison of our proposed method when trained with different dataset sizes. Samples means the number of training samples used.}
		\label{tab:ApiQ_saliency_aware_weight_preservation_reg_dataset_size}
		\resizebox{\textwidth}{!}{%
			\begin{tabular}{c c c c c c c c c c}
				\toprule
				\multirow{2}{*}{\textbf{Method}} 
				& \multirow{2}{*}{\textbf{Samples}}
				& \multicolumn{2}{c}{\textbf{PPL}\,$\downarrow$}
				& \multicolumn{6}{c}{\textbf{Accuracy (\%)}\,$\uparrow$} \\
				\cmidrule(lr){3-4}
				\cmidrule(lr){5-10}
				& 
				& \textbf{WikiText2} & \textbf{C4}
				& \textbf{WinoGrande} & \textbf{HellaSwag} 
				& \textbf{ArcC}    & \textbf{ArcE} 
				& \textbf{PiQA}    & \textbf{Average} \\
				\midrule
				\textbf{Full-Precision Model}      & - & 5.47 & 6.97 & 69.22 & 57.16 & 43.52 & 76.26 & 78.07 & 64.85 \\
				\midrule
				\textbf{Original ApiQ}          & 128 & 7.56 & 10.42 & 62.83 & 46.53 & 31.57 & 66.54 & 72.09 & 55.91 \\
				\textbf{ApiQ with Proposed Regularization}& 128 & 7.57 & 10.26 & 64.80 & 47.11 & 32.34 & 67.21 & 72.96 & 56.88 \\
				\midrule
				\textbf{Original ApiQ}          & 512 & 7.58 & 10.41  & 61.48 & 46.60 & 32.85 & 67.05 & 72.09 & 56.01 \\
				\textbf{ApiQ with Proposed Regularization}& 512 & 7.55 & 10.30 & 64.25 & 46.89 & 33.28 & 67.72 & 73.01 & 57.03 \\
				\bottomrule
			\end{tabular}
		}
	\end{table}
	
	\FloatBarrier
	
	The results are provided in Table \ref{tab:ApiQ_saliency_aware_weight_preservation_reg_dataset_size}. It can be seen that our method shows consistent improvement when the dataset size is increased and thus the performance of the baseline ApiQ is improved as well.

\end{document}